\documentclass[journal]{IEEEtran}

\usepackage{nicefrac}
\usepackage{rotating}
\usepackage{hyperref}
\usepackage{csquotes}
\usepackage[noadjust]{cite}
\usepackage{amsthm}
\usepackage{amsmath}           
\usepackage{amssymb}           
\usepackage[10pt]{moresize}
\usepackage{tikz}

\newcommand{\cfbox}[2]{%
\setlength{\fboxrule}{1.1pt}%
    \colorlet{currentcolor}{.}%
    {\color{#1}%
    \fbox{\color{currentcolor}#2}}%
}

\newcommand{\cfboxR}[2]{%
\setlength{\fboxrule}{1.1pt}%
    \colorlet{currentcolor}{.}%
    {\color{#1}%
    \fbox{\color{currentcolor}#2}}%
}

\newcommand{\cfboxRR}[2]{%
\setlength{\fboxrule}{1.25pt}%
    \colorlet{currentcolor}{.}%
    {\color{#1}%
    \fbox{\color{currentcolor}#2}}%
}

\newcommand{\cfboxRRR}[2]{%
\setlength{\fboxrule}{0.9pt}%
    \colorlet{currentcolor}{.}%
    {\color{#1}%
    \fbox{\color{currentcolor}#2}}%
}

\definecolor{redB}{rgb}{.75,0,0}
\definecolor{green1}{rgb}{.7,.7,.7}
\definecolor{blue1}{rgb}{.3,.3,1}
\definecolor{red1}{rgb}{.35,.65,.25}
\definecolor{red9}{rgb}{0.7075    0.4125    .85}
\definecolor{red11}{rgb}{0.2940    0.2100    0.5460}
\definecolor{red12}{rgb}{ 0.8000    0.4627    0.3961}
\definecolor{red10}{rgb}{0.2863    0.0471    0.7882}
\definecolor{other1b}{rgb}{0.8431    0.1882    0.1529}
\definecolor{other2b}{rgb}{0.9882    0.5529    0.3490}
\definecolor{other3b}{rgb}{0.9961    0.8784    0.5451}
\definecolor{other4b}{rgb}{1.0000    1.0000    0.7490}
\definecolor{other5b}{rgb}{0.8510    0.9373    0.5451}
\definecolor{other6b}{rgb}{0.5686    0.8118    0.3765}
\definecolor{other7b}{rgb}{0.1020    0.5961    0.3137}
\definecolor{other1c}{rgb}{0.759    0.169    0.138}
\definecolor{other2c}{rgb}{0.889    0.498    0.314}
\definecolor{other3c}{rgb}{0.896    0.791    0.491}
\definecolor{other4c}{rgb}{0.900    0.900    0.674}
\definecolor{other5c}{rgb}{0.766    0.844    0.491}
\definecolor{other6c}{rgb}{0.512    0.731    0.339}
\definecolor{other7c}{rgb}{0.092    0.536    0.282}
\definecolor{other1}{rgb}{0.6745    0.1506    0.1224}
\definecolor{other2}{rgb}{0.7906    0.4424    0.2792}
\definecolor{other3}{rgb}{0.7969    0.7027    0.4361}
\definecolor{other4}{rgb}{0.8000    0.8000    0.5992}
\definecolor{other5}{rgb}{0.6808    0.7498    0.4361}
\definecolor{other6}{rgb}{0.4549    0.6494    0.3012}
\definecolor{other7}{rgb}{0.0816    0.4769    0.2510}
\definecolor{other9}{rgb}{0.1506 0.1224 0.6745 }
\definecolor{other8}{rgb}{0.06    0.4    0.2}
\definecolor{red3}{rgb}{.20,.45,.15}
\definecolor{black}{rgb}{0,0,0}
\definecolor{green2}{rgb}{.45,.05,.25}
\definecolor{violeta1}{rgb}{.85,.35,.25}

\definecolor{red11a}{rgb}{0.3    0    0.6}
\definecolor{red11b}{rgb}{0    0    0.8}

\definecolor{other1d}{rgb}{0.9156    0.3705    0.2510}
\definecolor{other2d}{rgb}{0.9882    0.5529    0.3490}
\definecolor{other3d}{rgb}{0.9961    0.8784    0.5451}
\definecolor{other4d}{rgb}{1.0000    1.0000    0.7490}
\definecolor{other5d}{rgb}{0.8510    0.9373    0.5451}
\definecolor{other6d}{rgb}{0.5686    0.8118    0.3765}
\definecolor{other7d}{rgb}{0.1020    0.5961    0.3137}

\newcommand{\bx}{\mathbf{x}}

\usepackage[vlined,linesnumbered,ruled]{algorithm2e}
\SetKwInOut{Input}{$\,\,\,\,\,\,$Input}
\SetKwInOut{Output}{Output}
\SetKwComment{Comment}{}{}

\SetCommentSty{mycmtsty}

\begin{document}

\title{Hand-held Video Deblurring via \\ Efficient Fourier Aggregation}

\author{Mauricio~Delbracio 
        and~Guillermo~Sapiro,~\IEEEmembership{Fellow,~IEEE}
\IEEEcompsocitemizethanks{
\IEEEcompsocthanksitem This work was partially funded by: 
ONR, ARO, NSF, NGA, and AFOSR.
\IEEEcompsocthanksitem The authors are with the
Department of Electrical and Computer Engineering at Duke University.

e-mail: \{mauricio.delbracio, guillermo.sapiro\}@duke.edu

}
}

\maketitle

\begin{abstract}
Videos captured with hand-held cameras often suffer from a significant amount of blur, mainly 
caused by the inevitable natural tremor of the photographer's hand. 
In this work, we present an algorithm that removes blur due to camera shake by combining information
in the Fourier domain from nearby frames in a video.
The dynamic nature of typical videos with the presence of multiple moving objects and occlusions
makes this problem  of camera shake removal extremely challenging, in particular when low complexity is needed.
Given an input video frame, we first create a consistent registered version of temporally 
adjacent frames. Then, the set of consistently registered frames is block-wise fused in
the Fourier domain with weights depending on the Fourier spectrum magnitude.
The method is motivated from the physiological fact that camera shake blur has a random 
nature and therefore, nearby video frames are generally blurred differently. 
Experiments with numerous videos recorded in the wild, along with extensive comparisons, show that 
the proposed algorithm achieves state-of-the-art results while at the same time being much 
faster than its competitors.

\end{abstract}

\begin{IEEEkeywords}
Video deblurring, camera shake, Fourier accumulation
\end{IEEEkeywords}

\IEEEpeerreviewmaketitle

\section{Introduction}
\label{sec:introduction}

Videos captured with hand-held cameras often suffer from a significant amount of blur, 
mainly  caused by the tremor of the photographer hands. This problem is exacerbated 
when  shooting in dim light conditions because significant noise is introduced on top 
of the blur.   Although recent state-of-the-art optical image stabilizers mitigate this problem, 
their performance is far from being perfect.

The  acquisition of a video frame is traditionally modeled as a convolution,
\begin{equation}
v = u \star k + n,
\label{eq:convolution}
\end{equation}
where $v$ is the noisy and blurred observation, $u$ is the  underlying sharp image, $k$ is 
an unknown blurring kernel and $n$ is additive white noise. 
Blur in video frames can be caused by different phenomena.  All digital cameras will 
have a minimum amount of image blur given by the light integration on the camera 
sensor and the light diffraction on the camera aperture. In addition, image blur can 
be  consequence of wrongly setting the camera focus or having a finite depth of field. 
The presence of relative motion between the camera and the objects in the scene during
the frame acquisition will also result in blur. 
However, in many situations, when shooting 
with a hand-held  camera, the dominant contribution to the blur kernel is due the camera 
shake --caused by natural hand tremor.

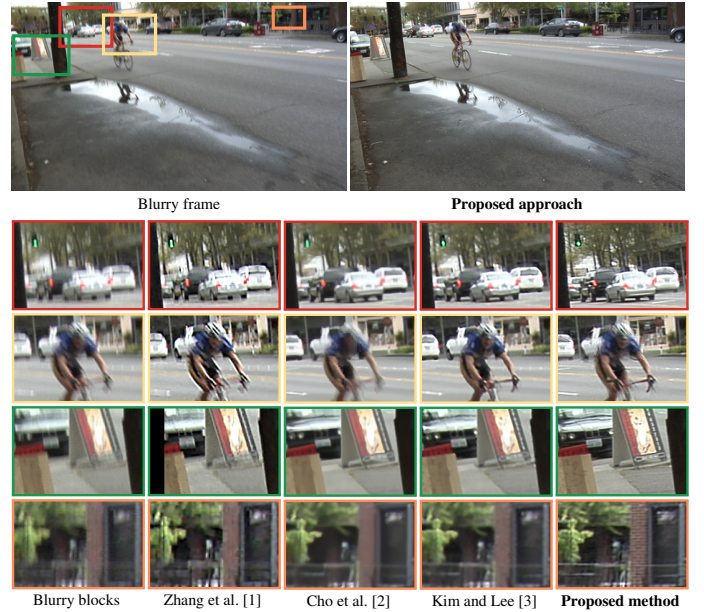
\begin{figure}[t]

\begin{minipage}[c]{.5\textwidth}
\ssmall

 \begin{minipage}[c]{.49\textwidth}
 \centering
 
\begin{tikzpicture}
    \node[anchor=north west,inner sep=0] (image) at (0,0) {\includegraphics[width=\textwidth,clip=true,trim=0 0 0 0]{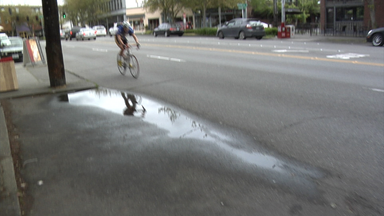}};
    \begin{scope}[x={(image.north east)},y={(image.south west)}]
        \draw[other1b, very thick] (0.145,0.03) rectangle (0.3,0.235); 
        \draw[other3b, very thick] (0.275,0.07) rectangle (0.43,0.275); 
        \draw[other7b, very thick] (0.007,0.18) rectangle (0.172,0.385); 
        \draw[other2b, very thick] (0.7812,0.0278) rectangle (0.8750,0.1389); 
    \end{scope}
\end{tikzpicture}

Blurry frame 
\end{minipage}
 \begin{minipage}[c]{.49\textwidth}
 \centering
 
\includegraphics[width=\textwidth,clip=true,trim=0 0 0 0]{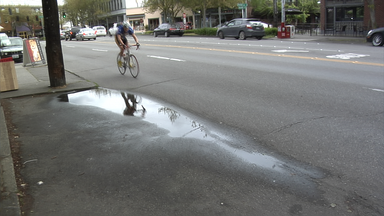}
 
\textbf{Proposed approach} 
 \end{minipage}

\vspace{.5em}

\begin{minipage}[c]{0.19\textwidth}
\centering
\cfboxRRR{other1b}{\includegraphics[width=.99\textwidth]{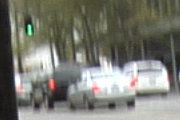}} \vspace{-.84em}

\cfboxRRR{other3b}{\includegraphics[width=.99\textwidth]{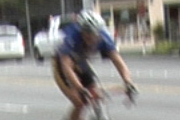}} \vspace{-.84em}

\cfboxRRR{other7b}{\includegraphics[width=.99\textwidth]{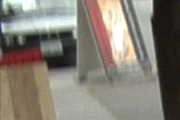}} \vspace{-.84em}

\cfboxRRR{other2b}{\includegraphics[width=.99\textwidth]{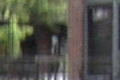}}

Blurry blocks
\end{minipage}
\hspace{-.1em}
%
%
%
%
%
\begin{minipage}[c]{0.19\textwidth}
\centering

\cfboxRRR{other1b}{\includegraphics[width=.99\textwidth]{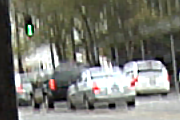}} \vspace{-.84em}

\cfboxRRR{other3b}{\includegraphics[width=.99\textwidth]{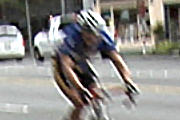}} \vspace{-.84em}

\cfboxRRR{other7b}{\includegraphics[width=.99\textwidth]{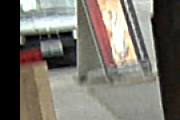}} \vspace{-.84em}

\cfboxRRR{other2b}{\includegraphics[width=.99\textwidth]{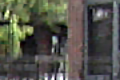}}

Zhang et al.~\cite{zhang2013multi}

\end{minipage}
\hspace{-.1em}
\begin{minipage}[c]{0.19\textwidth}
\centering

\cfboxRRR{other1b}{\includegraphics[width=.99\textwidth]{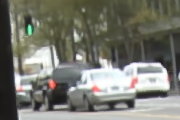}} \vspace{-.84em}

\cfboxRRR{other3b}{\includegraphics[width=.99\textwidth]{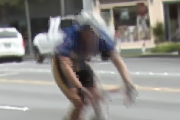}} \vspace{-.84em}

\cfboxRRR{other7b}{\includegraphics[width=.99\textwidth]{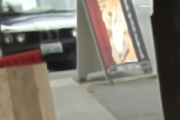}} \vspace{-.84em}

\cfboxRRR{other2b}{\includegraphics[width=.99\textwidth]{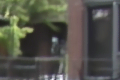}}

Cho et al.~\cite{cho2012video}

\end{minipage}
\hspace{-.1em}
\begin{minipage}[c]{0.19\textwidth}
\centering

\cfboxRRR{other1b}{\includegraphics[width=.99\textwidth]{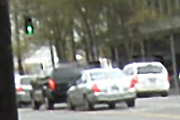}} \vspace{-.84em}

\cfboxRRR{other3b}{\includegraphics[width=.99\textwidth]{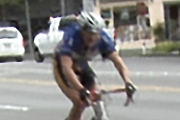}} \vspace{-.84em}

\cfboxRRR{other7b}{\includegraphics[width=.99\textwidth]{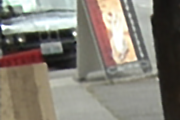}} \vspace{-.84em}

\cfboxRRR{other2b}{\includegraphics[width=.99\textwidth]{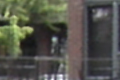}}

Kim and Lee~\cite{kim2015cvpr}

\end{minipage}
\hspace{-.1em}
\begin{minipage}[c]{0.19\textwidth}
\centering

\cfboxRRR{other1b}{\includegraphics[width=.99\textwidth]{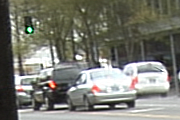}} \vspace{-.84em}

\cfboxRRR{other3b}{\includegraphics[width=.99\textwidth]{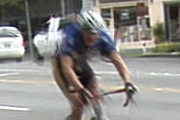}} \vspace{-.84em}

\cfboxRRR{other7b}{\includegraphics[width=.99\textwidth]{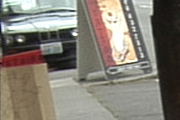}} \vspace{-.84em}

\cfboxRRR{other2b}{\includegraphics[width=.99\textwidth]{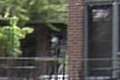}}

\textbf{Proposed method}
\end{minipage}

\end{minipage}
\caption{Video blur due to camera shake can be efficiently eliminated by aggregating
 information from nearby frames in the Fourier domain. Given an input (blurry) frame, 
 the proposed algorithm boosts its quality by performing a weighted local Fourier average 
 of aligned temporal neighboring frames. The proposed approach outperforms 
 the state-of-the-art while at the same time being significantly faster. See supplementary video for multiple additional results.}
\end{figure}

\begin{figure*}
\centering
\ssmall
\begin{minipage}[c]{0.068\textwidth}
\centering
1 \vspace{.1em}

\includegraphics[width=\textwidth]{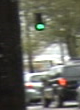}
\end{minipage}
\begin{minipage}[c]{0.068\textwidth}
\centering
2\vspace{.1em}

\includegraphics[width=\textwidth]{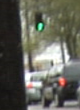}
\end{minipage}
\begin{minipage}[c]{0.068\textwidth}
\centering
3\vspace{.1em}

\includegraphics[width=\textwidth]{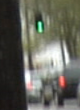}
\end{minipage}
\begin{minipage}[c]{0.068\textwidth}
\centering
4 \vspace{.1em}

\includegraphics[width=\textwidth]{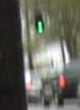}
\end{minipage}
\begin{minipage}[c]{0.068\textwidth}
\centering
5 \vspace{.1em}

\includegraphics[width=\textwidth]{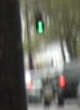}
\end{minipage}
\begin{minipage}[c]{0.068\textwidth}
\centering
 6\vspace{.1em}

\includegraphics[width=\textwidth]{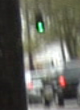}
\end{minipage}
\begin{minipage}[c]{0.068\textwidth}
\centering
7\vspace{.1em}

\includegraphics[width=\textwidth]{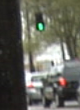}
\end{minipage}
\begin{minipage}[c]{0.068\textwidth}
\centering
 8\vspace{.1em}

\includegraphics[width=\textwidth]{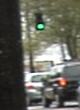}
\end{minipage}
\begin{minipage}[c]{0.068\textwidth}
\centering
9\vspace{.1em}

\includegraphics[width=\textwidth]{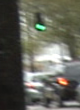}
\end{minipage}
\begin{minipage}[c]{0.068\textwidth}
\centering
10\vspace{.1em}

\includegraphics[width=\textwidth]{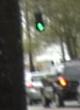}
\end{minipage}
\begin{minipage}[c]{0.068\textwidth}
\centering
11\vspace{.1em}

\includegraphics[width=\textwidth]{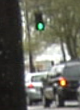}
\end{minipage}
\begin{minipage}[c]{0.068\textwidth}
\centering
12\vspace{.1em}

\includegraphics[width=\textwidth]{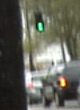}
\end{minipage}
\begin{minipage}[c]{0.068\textwidth}
\centering
13\vspace{.1em}

\includegraphics[width=\textwidth]{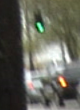}
\end{minipage}
\begin{minipage}[c]{0.068\textwidth}
\centering
14\vspace{.1em}

\includegraphics[width=\textwidth]{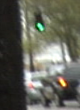}
\end{minipage}

\caption{Videos captured with hand-held cameras often contain blur. Due to the random 
nature of hand tremor, handshake blur is different in different frames of the video.}
\label{fig:videoBlur}
\end{figure*}

The classical deblurring mathematical formulation as an inverse deconvolution problem, seeks to
jointly estimate the camera motion path (or directly the blurring operator) and the underlying sharp image.  
Although this can produce good results~\cite{zhang2013multi}, it requires significant computational resources
 and it is very sensitive to a highly precise estimation of the camera motion path (or directly the blurring operator).
Other type of approaches rely on the detection of sharp key frames/regions and the propagation of these 
to restore the blurry ones. Methods of this type are based on the existence and the detection of lucky 
frames or lucky regions, i.e., parts of the blurry image appearing sharp in  other frames. The goal is then to
interpolate those lucky frames/regions to substitute the unlucky blurry ones.
These approaches exploit the fact that the camera shake originated from the photographer's hand 
tremor is essentially random~\cite{xiao2006camera,carignan2010quantifying,gavant2011physiological}. 
This implies that, in general, the camera movements in different video frames are independent, leading to 
different image blurs and the existence of (potentially less blurred) lucky frames. An example of this is shown in Figure~\ref{fig:videoBlur}.

In~\cite{delbracio2015cvpr}, we presented an algorithm that combines an image 
burst by creating a new image whose Fourier  spectrum takes, for each frequency, 
the value with the largest Fourier magnitude in the burst. Similar ideas were also explored 
in the context of astronomical imaging  through atmospheric turbulence~\cite{garrel2012highly}. 
Since each image in the burst is blurred differently, and that blurring acts as a low pass filter, 
the reconstructed image picks what is less attenuated, in the Fourier domain,  from each 
image of the burst. This algorithm produces state-of-the art results on bursts capturing 
static scenes and is significantly faster than those based on deconvolution ideas or classical
lucky imaging techniques~\cite{delbracio2015tip}.

In this paper we take on these ideas to restore blurry videos caused by camera shake. 
In a typical case, and contrary to the static scene case, the frame fusion is non-trivial
due to the  dynamic nature of the scene with the presence of multiple moving objects and occlusions.
This problem  has strong requirements not only from the
video quality perspective but also regarding processing time and
memory consumption. 

Instead of introducing a complex model of the blurring and the camera
motion in the sequence (e.g., requiring different motion layers,
object segmentation and an accurate forward model), we propose to
deblur each frame of the sequence by locally fusing the consistent
information present in nearby frames. 
%
%
%
%
Since the vast majority of consumer hand-held videos are aimed at capturing dynamic scenes,
this is extremely challenging, in particular at low cost.
 Specifically, given a frame (reference) and its nearby
ones, the proposed algorithm first \emph{consistently} registers these
frames to the reference, and then locally applies the weighted Fourier
fusion. The consistent registration produces a new equivalent set of
frames that {\it locally} has the same spectrum as the reference up to the
effects of a blurring kernel. This enables us to locally apply the
Fourier fusing scheme~\cite{delbracio2015cvpr} with limited to no artifacts. 
The consistent registration and the local Fourier fusion are what make the
algorithm very efficient in terms of computational resources.
This procedure yields results (1) without 
image blur and (2) with a significantly reduced amount of noise, due to 
the aggregation of different frames.

The presented evaluation in many real video sequences shows that the
video quality is significantly improved. A detailed comparison to 
state-of-the-art video deblurring algorithms shows that the proposed 
approach produces similar or better results while being significantly faster, 
in particular due to the avoidance of explicit kernel computation and 
deconvolution.

The remainder of the paper is organized as follows.
In Section 2 we discuss the closely related work, while in Section 3 
we explain the principles ruling the proposed camera shake video 
removal and the corresponding mathematical framework. In Section 4, 
we present and discuss the proposed video deblurring algorithm while in 
Section 5 we present results in real data. We finally close in Section 6 
providing the final conclusions, some limitations, and several ideas regarding 
future work.

\section{Related Work}
\label{sec:relatedWork}
A thorough analysis of image/video deblurring is far beyond the scope of the present work. 
As aforementioned, image blur may have multiple causes. 
For instance, blur caused by the fast movement of objects presents 
very different characteristics than camera shake blur. 
The hypotheses of randomness and independence in successive frames, reasonable assumption for camera shake, 
does not hold in general for to the movement of objects in the scene (which usually keep  the same movement along several frames). 
In this work, we focus exclusively on the removal of blur  due to the random camera movement.  
Therefore, we will not delve in the vast existent literature that concentrates on removing  
object motion blur (see e.g., \cite{levin2006blind,kim2013dynamic,kim2014segmentation}) and 
we concentrate on general deblurring techniques that target (or can be easily adapted to) camera shake removal.

For what follows, it is enough to note that there are mainly two different kinds of approaches to reduce camera shake blur in videos. 
The first one formulates the deblurring problem as an inverse problem (e.g., deconvolution), while the second one seeks to detect 
and transfer (or aggregate) the sharp information from all the frames to produce a sharper sequence. 

\vspace{.5em}

\noindent \textbf{Deblurring as an inverse problem.}
In recent years, many successful image restoration algorithms, which try to blindly recover the underlying sharp image, have emerged. 
Most of these works combine natural image priors, assumptions on the blurring operator or the camera path, and sophisticated 
optimization algorithms, to simultaneously solve an inverse estimation problem for recovering both the blurring kernel and 
the sharp image e.g.,~\cite{fergus2006removing,shan2008high,cai2009blindsingle,cho2009fast,krishnan2011blind,xu2013unnatural,schelten2014localized,michaeli2014blind}.

Due to its small spatial support, the blurring kernel estimation is an easier problem to solve than simultaneously estimating both the kernel and the 
sharp image~\cite{levin2009understanding,levin2011efficient}. However, even in non-blind deconvolution, i.e., when the blurring kernels are known, 
the problem is  generally ill-posed, because the blur introduces zeros in the frequency domain, which hinders the estimation.

Video deblurring is very related to multi-image blind deconvolution e.g.,~\cite{chen2008robust,cai2009blind,sroubek2012robust,zhu2012deconvolving,zhang2013multi}. 
Cai et al.~\cite{cai2009blind} showed that given multiple observations, the sparsity of the image under 
a tight frame is a good measurement of the clearness of the recovered image. Having access to multiple input blurry images improves the accuracy of 
identifying the motion blur kernels and reduces the illposedness of the problem.
Rav-Acha and Peleg~\cite{rav2005two} stated that \emph{``two motion-blurred images are better than one,''}  whenever the motion 
directions are different.
Most of multi-image deconvolution algorithms introduce cross-blur penalties between each pair of input images. 
This has the problem of growing combinatorially with the number of considered images. 
Zhang et al.~\cite{zhang2013multi} proposed a Bayesian framework for coupling all the unknown blurring kernels and the latent sharp image in 
a unique prior. Although this formulation produces in general good looking sharp images, its optimization is very slow and may require several 
minutes for filtering  a high-definition (HD) frame using its nearby ones. In addition, virtually all multi-image deconvolution algorithms require 
that all the input images are aligned and that the content is the same (static scene).

Li et al.~\cite{li2010generating} propose to estimate the camera motion and to explicitly model the video blur as a function of the motion being estimated. 
They formulate and optimize a joint energy function between the underlying sharp sequence and motion parameters. In~\cite{paramanand2013non}, the authors propose a method to estimate the latent sharp image of a bilayer static scene using two motion blurred observations.
This was extended to a more general case having layers with different motions in \cite{wulff2014modeling}.

Very recently, Kim and Lee \cite{kim2015cvpr} proposed to simultaneously tackle the problem of optical flow estimation and frame restoration 
in general blurred videos. This is done by simultaneously estimating the optical flow and latent sharp frames through the minimization of a single non-convex energy function. Addressing these two problems simultaneously requires a much more complex optimization, 
due to the more sophisticated forward model linking all the blurry observations.

%
%

%
All these works propose to solve an inverse problem of image restoration (e.g., deconvolution). 
The main drawback of this approach, on top of the computational burden, 
is that if the forward model is not accurate (or it is not accurately estimated), 
the restored sequence will contain strong artifacts (such as ringing). This is often observed in all the mentioned algorithms.
\vspace{.5em}

\noindent \textbf{Deblurring by transferring sharp information.}
A popular technique in astronomical photography, known as \emph{lucky imaging} or \emph{lucky exposures}, 
is to take a series of thousands of short-exposure images and then select and fuse only the top sharpest ones~\cite{law2006lucky}.  
Fried~\cite{fried1978probability} mathematically showed that with high probability one will  capture a sharp lucky exposure if  the captured video is long enough.
 Astronomical lucky frame selection methods are based on the brightness of the  brightest speckle~\cite{law2006lucky}.  Others propose 
 to measure the local sharpness from the energy of the gradient or the image 
Laplacian~\cite{john2005multiframe,aubailly2009automated,joshi2010seeing,haro2012photographing}.
Classical lucky imaging methods try to generate a single image from a static video (or multiple frames) instead of restoring the full video.

To get rid of shaky motion frames in videos, Matsushita et al.~\cite{matsushita2006full} propose to transfer, by interpolation, sharp image pixels  
from nearby  frames to increase the sharpness of the blurry ones. In a similar fashion as lucky imaging techniques, these transfer-type algorithms
 are based on the observation that due to the random nature of camera shake, not all video frames are equally blurred.
To achieve deblurring, they propose a motion inpaiting algorithm that enforces
spatial and temporal consistency in static and dynamic image regions.
The main drawback is that camera motion is modeled and estimated  by pure homographies;  thus,
in many practical scenarios this model is not accurate and leads to visual artifacts and below-par image quality.

Similar ideas were explored by Cho et al.~\cite{cho2012video}, where the authors propose to replace blurry patches with a linear combination of 
similar but sharper ones from nearby frames.
A rough estimation of the blurring kernels is used to detect the most similar patches in nearby frames. Then, each patch is replaced by a weighted 
average of the similar ones. The weights are a combination of the similarity between the patches, and a luckiness term that gives more weight to patches
 that are detected as potentially  sharper. Although this algorithm produces in general good results, it sometimes tends to over-smooth the image due to
  the non-local average of patches. In the results section we show a detailed comparison to this method.

A general disadvantage of  traditional lucky imaging approaches is that they only rely on  sharpness measures 
and do not exploit the fact that camera shake blur occurs in different directions in different frames.

Garrel et al.~\cite{garrel2012highly} introduced a selection scheme for astronomic images, based on the relative 
strength of signal for each Fourier frequency.  Similarly, in~\cite{delbracio2015cvpr,delbracio2015tip}, the Fourier Burst Accumulation (\textsc{fba}) algorithm 
fusions an image burst by creating a new image whose Fourier spectrum takes for each frequency the value having the largest Fourier magnitude in the burst. 
These procedures make a much more efficient use of the complimentary information contained in each blurred frame.

\section{Removing Blur in Hand-held Cameras}
Videos captured using hand-held cameras often contain image blur which significantly damages the overall quality.
Typical blur sources can be separated into those mainly depending on the scene (e.g., objects moving, depth-of-field),
and those depending on the camera and the movement of the camera (camera shake, autofocusing).

Image blur due to camera shake can be visually very disturbing.  Fortunately, in many cases, this blur is temporal, 
non-stationary and of rapid change. This implies that, in general, the blur due to camera shake in each frame will be different
from the blur in nearby frames. In this work, we propose an algorithm that exploits this phenomenon by 
aggregating information from nearby frames to improve the quality of every frame in the video sequence.
The proposed algorithm is inspired on the Fourier deblurring fusion introduced in
\cite{delbracio2015cvpr,garrel2012highly}. 
Let us point out that going from a static-scene multiimage deblurring 
algorithm to an algorithm for removing camera shake blur in dynamic videos, while keeping 
the simplicity and complexity low, is extremely challenging.  This is the reason 
why, in general,  multi-image deblurring algorithms have not been (yet)
successfully extended to remove camera shake blur in real dynamic videos. 
In what follows, we briefly describe the main ideas behind 
these approaches and the mathematical formalism.
 
\subsection*{The Weighted Fourier Accumulation Principle}

Let $u$ be a digital image (e.g., a video frame) defined in a regular grid indexed by the $2D$ position $\bx$.
Let  $\mathcal F$ denote the Fourier Transform and $\hat{u}$ the Fourier Transform of $u$.  
The Fourier domain is indexed by the $2D$ frequency $\zeta$.
We will assume, without loss of generality, that the kernel $k$,  comprising all blur sources, is normalized such that
$\int k(\bx) d\bx = 1$. The blurring kernel is nonnegative since the integration of incoherent light is always nonnegative. 
This implies that the camera blur acts as a low pass filter and never amplifies the Fourier spectrum (that is, $\forall \zeta, |\hat{k}(\zeta) | \le 1$, see~\cite{delbracio2015cvpr}).

Let us assume first that we have access to a video sequence of $2M+1$ consecutive frames centered 
at the reference frame $v_0$ ($M$ frames preceding the reference, the reference, and $M$ frames succeeding the reference),
\begin{align}
v_i = (u \circ \tau_i) \star k_i + n_i + o_i, \quad \text{for} \quad i=-M,\ldots,M,
\label{eq:burstComplete}
\end{align}
where $u$ is the latent sharp reference image, $k_i$ is the blurring kernel affecting the frame $i$, $n_i$ noise in the capture,  $o_i$ models the parts of the frame that are different from the reference scene
(e.g., occlusions), and $\tau_i$ models the geometric transformation between the frame $i$ and the reference ($\tau_0$ is  the identity function).

The rationale behind the \textsc{fba} algorithm developed for still-bursts~\cite{delbracio2015cvpr} is that  since blurring kernels do not amplify the Fourier spectrum, the reconstructed image should pick from each image of the burst what is less attenuated in the Fourier domain.
 
The principle for still images assumes that all the captured images are equal up to the effect of a shift invariant blurring kernel and additive noise, i.e.,
\begin{align}
v_i = u  \star k_i + n_i, \quad \text{for} \quad i=-M,\ldots,M.
\label{eq:burst}
\end{align}
Let $p$ be a non-negative integer,  and $\{v_i\}$ be a set of aligned images of a static scene (given by Eq.~\eqref{eq:burst}),
then the \textsc{fba} average is given by 
\begin{align}
{\bar{u}} =  \mathcal{F}^{-1} \left( \sum_{i=-M}^M w_i (\zeta) \cdot \hat{v}_i (\zeta) \right),  \,\, w_i(\zeta)   = \frac{ |\hat{v}_i (\zeta) |^p }{\sum_{j=1}^M | \hat{v}_j (\zeta)|^p},
\label{eq:fourierWeightsOrig}
\end{align}
where $\hat{v}_i(\zeta)$ is the Fourier Transform of the individual image $v_i(\bx)$. 
The Fourier weight $w_i(\zeta)$ controls the contribution of the frequency $\zeta$ of image $v_i$ to the final reconstruction $\bar{u}$.  Given the Fourier frequency $\zeta$, for $p>0$, the larger the value of   $|\hat{v}_i (\zeta)|$, the more $\hat{v}_i (\zeta)$ contributes  to the average, reflecting the fact that the strongest frequency values represent the least attenuated components. Note that this is not the result of assumed image models, but a direct consequence of the standard image formation model (3) and the physiology of hand tremor.

The parameter $p$ controls the behavior of the Fourier aggregation. 
If $p=0$, the restored image is just the arithmetic average of the burst, while if $p\to \infty$, each reconstructed frequency takes the maximum value of that frequency  along the burst. 

While this extremely simple algorithm produces very good (state-of-the-art) results in the case of static scenes, it cannot be directly applied to restore general hand-held videos. In a typical video sequence, there are moving objects, occlusions, and changes of illumination, that need to be considered.  In what follows we describe how we can incorporate these dynamic components into the ideas behind the \textsc{fba} algorithm to deal with real videos.

\section{Video Deblurring: Algorithm Overview}

\begin{figure*}[tpb]

\begin{minipage}[c]{0.160\textwidth}
\centering
\scriptsize Input Frame  $v_i (\bx)$ \vspace{.1em}

\includegraphics[width=\textwidth]{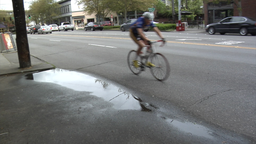} \vspace{-.9em}

\cfbox{other7b}{\includegraphics[width=.46\textwidth]{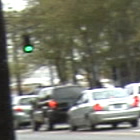}} \hspace{-.3em}
\cfbox{other1b}{\includegraphics[width=.46\textwidth]{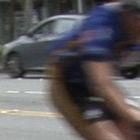}}

\end{minipage}
\begin{minipage}[c]{0.160\textwidth}
\centering
\scriptsize Frame Warping  $v_i \circ \tau_i (\bx)$ \vspace{.1em}

\includegraphics[width=\textwidth]{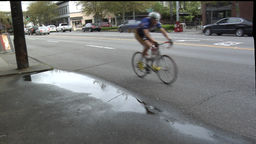}  \vspace{-.9em}

\cfbox{other7b}{\includegraphics[width=.46\textwidth]{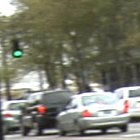}} \hspace{-.3em}
\cfbox{other1b}{\includegraphics[width=.46\textwidth]{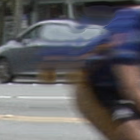}}

\end{minipage}
\begin{minipage}[c]{0.160\textwidth}
\centering
\scriptsize Cons. Map  $\text{cMap}_i(\bx)$  \vspace{.1em}

\includegraphics[width=\textwidth]{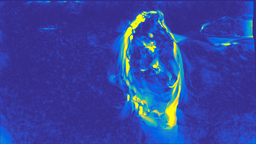}   \vspace{-.9em}

\cfbox{other7b}{\includegraphics[width=.46\textwidth]{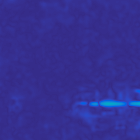}} \hspace{-.3em}
\cfbox{other1b}{\includegraphics[width=.46\textwidth]{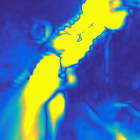}}

\end{minipage}
\begin{minipage}[c]{0.160\textwidth}
\centering
\scriptsize Cons. Mask  $\text{M}_i(\bx)$ \vspace{.1em}

\includegraphics[width=\textwidth]{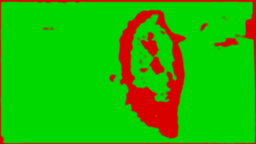}   \vspace{-.9em}

\cfbox{other7b}{\includegraphics[width=.46\textwidth]{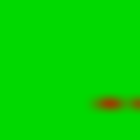}} \hspace{-.3em}
\cfbox{other1b}{\includegraphics[width=.46\textwidth]{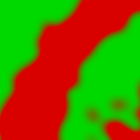}}

\end{minipage}
\begin{minipage}[c]{0.162\textwidth}
\centering
\scriptsize Cons. Registration $v^0_i (\bx)$ \vspace{.1em}

\includegraphics[width=\textwidth]{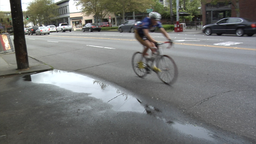}   \vspace{-.9em}

\cfbox{other7b}{\includegraphics[width=.46\textwidth]{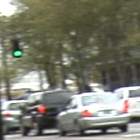}} \hspace{-.3em}
\cfbox{other1b}{\includegraphics[width=.46\textwidth]{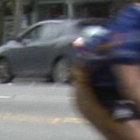}}

\end{minipage}
\begin{minipage}[c]{0.160\textwidth}
\centering
\scriptsize Reference Frame $v_0 (\bx)$ \vspace{.1em}

\includegraphics[width=\textwidth]{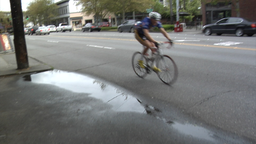}   \vspace{-.9em}

\cfbox{other7b}{\includegraphics[width=.46\textwidth]{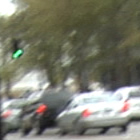}}  \hspace{-.3em}
\cfbox{other1b}{\includegraphics[width=.46\textwidth]{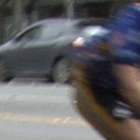}}

\end{minipage}

\caption{Consistent registration example. In this video sequence there are several moving objects (in particular the biker) that hinder the image registration. As shown in the image crops, there are some image regions that can be easily mapped from one frame to the other and some that cannot. The interpolated frame in the second column clearly shows that the biker is wrongly interpolated, being mixed with the car behind him. The computed consistency map  prevents these pixels from being interpolated as shown in the consistent registration $v^0_i$. From the two crops, there is one that is successfully interpolated (green) while the other is mostly copied from the reference (red).}
\label{fig:consistent_registration}
\end{figure*}

Given a reference input blurry frame and its $2M$ preceding/succeeding frames, our goal is to generate a new version
of the reference image having less noise and less blur. To that aim,  we proceed by (i) consistently registering the
adjacent 2M frames to the reference one, and then (ii)  locally aggregating the registered frames with a local
extension of \textsc{fba}.

The goal of step (i) is to generate an \emph{equivalent} input image sequence  that is aligned in a way that
each frame appears the same as the reference (up to a local difference in blur and noise). This enables in
step (ii) the local application of the \textsc{fba} procedure, without introducing artifacts. In the following, we detail both key components.

\subsection{Consistent Frame Registration}

Estimating the motion from a sequence of images is a longstanding problem in computer vision
 (see, for example, the general reviews of Barron et al.~\cite{barron1994performance} and Baker et al.~\cite{baker2011database}). 
The problem known as optical flow aims at computing the motion of each pixel  from consecutive frames. Most techniques tackle 
the problem from a variational perspective. Typically, the fitting (data) term assumes the conservation of some property 
(e.g., pixel brightness) along the sequence. A regularization term is then used to constraint the possible solutions, and 
to provide some regularity to the estimated motion field. There are many existing variants depending on the combination
 of fitting/regularization terms used~\cite{baker2011database}.

\vspace{.5em}

\noindent \textbf{Registration of temporal-variant blur sequences.}
In the general case where the video sequence is degraded by temporal-variant blur the problem of defining a \emph{correct}
 frame alignment is not well defined. 
In \cite{yuan2007blurred}, the authors present an effective algorithm for aligning a pair of blurred/non-blurred images using
a prior on the kernel sparseness. 
The method  seeks the best possible alignment (from a predefined set of rigid transformations) producing the sparsest 
kernel compatible with the blurry/sharp image pair. This algorithm requires that one of the images is sharp (the reference) 
which burdens its application in general videos. 
In addition, the amount of predefined possible rigid transformation reduces its application to videos of static scenes.
 
A more general idea of what constitutes  a correct alignment between differently blurred frames is introduced in~\cite{delbracio2015tip}.
An image sequence $\{v_i\}$ is said to be correctly aligned to the underlying sharp image $u$, if each $v_i$ satisfies 
$v_i = u \star k_i + n_i$,
where $n_i$ models random white noise and $k_i$ is a blurring kernel having vanishing first moment. 
%
This constraint on the blurring kernel implies that the kernel does not drift the image $u$, 
so each $v_i$ is aligned to~$u$ (see Appendix in~\cite{delbracio2015tip}).
Although this definition is more general than the previous one, it does not lead to
a (straightforward) construction of an optical flow estimation algorithm for 
blurred sequences. 

Recently, in~\cite{kim2015cvpr}, the authors propose to simultaneously estimate the optical flow and the latent sharp frames
by minimizing a non-convex function. The blurring operator is  assumed to be locally piecewise linear, and is determined by 
the optical flow. Since this problem is very ill-posed the method relies on strong spatial and temporal regularizations for both the
optical flows and the latent sharp images.
The cost of tackling these two problems simultaneously is a much more complex optimization, 
and a much more sophisticated forward model binding the blurry noisy observations.
 
As we detail in what follows, in this work, we proceed in a much simpler way. 
Nevertheless, we believe it is important to analyze how to address the optical
flow estimation when the sequence is perturbed by blur. This will be subject of future work.

One way of making more robust the computation of optical flow when image 
blur is present is by subsampling the input image sequence and computing the optical 
flow at a coarser scale. In this scenario, the impact of image blur is less significant. 
This brings up an obvious tradeoff between the optical flow resolution 
(and the corresponding alignment) and the level 
of blur to tolerate.

\vspace{.5em}

\noindent \textbf{Handling occlussions.}
Traditional optical flow estimation techniques do not generally yield symmetrical motion fields. 
Estimating the flow from one image to the next (forward estimation) generally does not yield the 
same result as estimating the flow in the opposite direction (backward estimation). 
The main reason for this is that many pixels get occluded when going from one frame to the other.

A  direct way of taking into account occlusions, is by jointly estimating forward and backwards optical flow. 
Alvarez et al.~\cite{alvarez2002symmetrical} exploited the fact that non-occluded pixels should have symmetric
forward and backward optical flows. 
A different appealing idea, given the fact that one has access to a complete video sequence, is to explicitly model the
detection of occlusions using more than two frames in a sequence (\cite{ince2008occlusion,ballester2012}). 
However, for simplicity, and to reduce the computational complexity of the algorithm, we opted to use only two 
frames and estimate the forward and backward flows independently and then cross-check them for consistency; see next.

\vspace{.5em}

\noindent \textbf{Consistent pixels.} 
Let $v_0$ be the reference image and $v_i$ one of the $i=-M,\ldots,M$ input frames that need to be registered to $v_0$. To apply the \textsc{fba} all the frames need to be the same
up to the effect of a centered shift invariant blur and noise. To satisfy these requirements, we first estimate the geometric transform between each frame and the reference, and then proceed to interpolate the set of consistent pixels (those that are in both frames and can be mapped through a  geometric transform). 

Let $\tau^0_i$ be an estimation of the optical flow from frame $v_i$ to the reference $v_0$, and similarly $\tau^i_0$ be the optical flow from the reference $v_0$ to $v_i$. 
Let $\text{cMap}_i(\bx)$ represent the inconsistency between the forward and backward optical flow estimation, that is,
\begin{align}
 \text{cMap}_i(\bx) := |( \tau^0_i \circ \tau^i_0 )(\bx)   - \bx|.
\end{align}
We consider a pixel $\bx$ to be consistently registered if $\text{cMap}_i(\bx)  \le \epsilon$, where $\epsilon$ is a given tolerance (in all the experiments $\epsilon=1$). 

Let $M_i$ be a mask function representing all the consistent pixels: $M_i(\bx) = 1$ if $\bx$ is consistent, and $0$ otherwise. 
Then, we create a new compatible version of $v_i$ by the following image blending
\begin{align}
v^0_i(\bx) =  M_i(\bx) \cdot ( v_i \circ \tau^0_i )(\bx) + (1-M_i(\bx)) \cdot v_0(\bx).
\label{Eq:blending}
\end{align}
This new frame $v^0_i$ propagates the \emph{reference-compatible} information present in the frame $v_i$ to the frame $v_0$ and keeps the reference values in the inconsistent area. The registered set has locally the same content as the reference, up to the effect of blur and noise. Note that even in the case that the frame $v_i$ was originally blurred with a shift invariant kernel, the warped frame $v^0_i$ might now be blurred with a shift variant blur due to the blending. This imposes the need to apply the Fourier fusion locally.

We post-process the mask $M_i(\bx)$ to avoid artifacts when doing the blending in~\eqref{Eq:blending}. The mask is first dilated, and then it is smoothed using a Gaussian filter to produce a smooth transition between both components. The details are given in Algorithm~\ref{algo:pAggregation} (lines 1--7). 

To compute the optical flow we used the algorithm from Zach et al.~\cite{hamprecht2007duality}, in particular the implementation given in~\cite{perez2013ipol}.  This algorithm is based on the minimization of an energy function containing a data fitting term using the $L_1$ norm, and a regularization term on the total variation of the motion field.  To accelerate the estimation and to mitigate the effects of blur, the optical flow is computed at 1/3 of the original resolution and then upsampled.

Figure~\ref{fig:consistent_registration} shows an example of the results of registering one image to the reference frame, in the presence of moving objects and occlusions. To avoid creating image artifacts, we take a conservative approach and discard difficult pixels. This is done at the expense of loosing potentially valuable information for the aggregation.

\subsection{Local Deblurring through Efficient Fourier Accumulation}

Due to the non-local nature of the Fourier decomposition, the Fourier aggregation in Eq.~\eqref{eq:fourierWeightsOrig} requires that the input images are uniformly blurred (shift invariant kernel).  The consistent registration previously described generates a set of $2M+1$ frames that locally have the same content up to the effect of a local blurring kernel and noise.  Thus, by splitting the frames into small blocks, the probability of satisfying the shift invariant blur assumption within each block is increased.  This  approximation is non critical to the final aggregation since the \textsc{fba} procedure does not force an inversion (or even computes the kernels), thus avoiding the creation of artifacts when the blurring model is not fully respected.

We split each registered image $v^0_i$ into a set of partially-overlapped blocks of  $b\times b$ pixels $\{P^l_i\}$ (position indexed by super-index $l=1,\ldots, n_l$), and then apply the \textsc{fba} procedure separately to each set of blocks.
Given the registered blocks $\{P^l_i\}_{i=-M}^M$ we directly compute the corresponding Fourier transforms  $\{\hat{P}^l_i\}_{i=-M}^M$.
To stabilize the Fourier weights, $|\hat{P}^l_i|$ is smoothed before computing the weights,
$
|\bar{\hat{P}}^l_i| =  G_\sigma  |\hat{P}^l_i|,
$
where $G_\sigma$ is a Gaussian filter of standard deviation $\sigma$.\footnote{The value of $\sigma$ controls the low pass filter and  was set to $\sigma =  \nicefrac{50}{b}$.}
Then, the Fourier fusion of the set of blocks is 
\begin{align}
P^l =  \mathcal{F}^{-1} \left ( \sum_{i=-M}^M w_i \cdot \hat{P}^l_i \right),  \qquad  w_i   = \frac{ |\bar{ \hat{P}}^l_i |^p }{\sum_{j=-M}^M | \bar{\hat{P}}^l_j |^p}.
\label{eq:fourierWeights}
\end{align}

Since blocks are partially-overlapped to mitigate boundary artifacts, in the end we have more than one estimate for each image pixel (e.g., a pixel belongs to up to 4 half-overlapped blocks). The final image is created by averaging  the different estimates coming from the overlapped blocks. The local Fourier fusion  is detailed in Algorithm~\ref{algo:pAggregation} (lines 8--22).

\begin{figure}[tbp]
\centering
\small

\begin{minipage}[c]{\columnwidth}
\centering

\begin{minipage}[c]{.235\textwidth}
\centering
\begin{tikzpicture}
        \node[anchor=north west, inner sep=0] (image) at (0,0) {\cfboxR{other1}{\includegraphics[width=.98\textwidth]{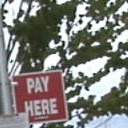}}};
        \node[anchor=north west,inner sep=0]  at (0.04,-0.04) { {\fboxsep 1pt \colorbox{white}{\textcolor{other1}{Ref-3}}}};
\end{tikzpicture}
\end{minipage}\hspace{.07em}
\begin{minipage}[c]{.235\textwidth}
\centering
\begin{tikzpicture}
        \node[anchor=north west, inner sep=0] (image) at (0,0) {\cfboxR{other2}{\includegraphics[width=.98\textwidth]{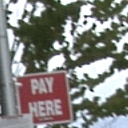}}};
        \node[anchor=north west,inner sep=0]  at (0.04,-0.04) { {\fboxsep 1pt \colorbox{white}{\textcolor{other2}{Ref-2}}}};
\end{tikzpicture}
\end{minipage}\hspace{.07em}
\begin{minipage}[c]{.235\textwidth}
\centering
\begin{tikzpicture}
        \node[anchor=north west, inner sep=0] (image) at (0,0) {\cfboxR{other3}{\includegraphics[width=.98\textwidth]{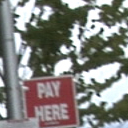}}};
        \node[anchor=north west,inner sep=0]  at (0.04,-0.04) { {\fboxsep 1pt \colorbox{white}{\textcolor{other3}{Ref-1}}}};
\end{tikzpicture}
\end{minipage}\hspace{.07em}
\begin{minipage}[c]{.235\textwidth}
\centering
\begin{tikzpicture}
        \node[anchor=north west, inner sep=0] (image) at (0,0) {\cfboxR{other4}{\includegraphics[width=.98\textwidth]{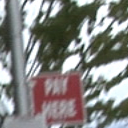}}};
        \node[anchor=north west,inner sep=0]  at (0.04,-0.04) { {\fboxsep 1pt \colorbox{white}{\textcolor{other4}{ Ref }}}};
\end{tikzpicture}
\end{minipage}

\vspace{.2em}

\begin{minipage}[c]{.235\textwidth}
\centering
\begin{tikzpicture}
        \node[anchor=north west, inner sep=0] (image) at (0,0) {\cfboxR{other5}{\includegraphics[width=.98\textwidth]{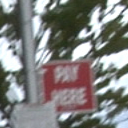}}};
        \node[anchor=north west,inner sep=0]  at (0.04,-0.04) { {\fboxsep 1pt \colorbox{white}{\textcolor{other5}{Ref+1}}}};
\end{tikzpicture}
\end{minipage}\hspace{.07em}
\begin{minipage}[c]{.235\textwidth}
\centering
\begin{tikzpicture}
        \node[anchor=north west, inner sep=0] (image) at (0,0) {\cfboxR{other6}{\includegraphics[width=.98\textwidth]{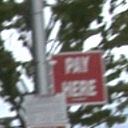}}};
        \node[anchor=north west,inner sep=0]  at (0.04,-0.04) { {\fboxsep 1pt \colorbox{white}{\textcolor{other6}{Ref+2}}}};
\end{tikzpicture}
\end{minipage}\hspace{.07em}
\begin{minipage}[c]{.235\textwidth}
\centering
\begin{tikzpicture}
        \node[anchor=north west, inner sep=0] (image) at (0,0) {\cfboxR{other7}{\includegraphics[width=.98\textwidth]{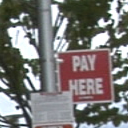}}};
        \node[anchor=north west,inner sep=0]  at (0.04,-0.04) { {\fboxsep 1pt \colorbox{white}{\textcolor{other7}{Ref+3}}}};
\end{tikzpicture}
\end{minipage}\hspace{.07em}
\begin{minipage}[c]{.235\textwidth}
\centering
\begin{tikzpicture}
        \node[anchor=north west, inner sep=0] (image) at (0,0) {\cfboxR{red12}{\includegraphics[width=.98\textwidth]{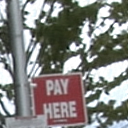}}};
        \node[anchor=north west,inner sep=0]  at (0.04,-0.04) { {\fboxsep 1pt \colorbox{white}{\textcolor{red12}{Output}}}};
\end{tikzpicture}
\end{minipage}

\end{minipage}

\vspace{1em}

\begin{minipage}[c]{\columnwidth}
\centering

\begin{minipage}[c]{.235\textwidth}
\centering
\begin{tikzpicture}
        \node[anchor=south west,inner sep=0] (image) at (0,0) {\cfboxR{other1}{\includegraphics[width=.98\textwidth]{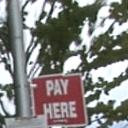}}};
        \begin{scope}[x={(image.south east)},y={(image.north west)}]
            \node[anchor=north east,inner sep=0] (image) at (1,1) {\cfboxR{other1}{\includegraphics[width=.35\textwidth]{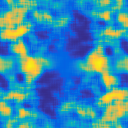}}};
        \end{scope}
    \end{tikzpicture}
\end{minipage}\hspace{.07em}
\begin{minipage}[c]{.235\textwidth}
\centering
\begin{tikzpicture}
        \node[anchor=south west,inner sep=0] (image) at (0,0) {\cfboxR{other2}{\includegraphics[width=.98\textwidth]{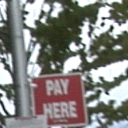}}};
        \begin{scope}[x={(image.south east)},y={(image.north west)}]
            \node[anchor=north east,inner sep=0] (image) at (1,1) {\cfboxR{other2}{\includegraphics[width=.35\textwidth]{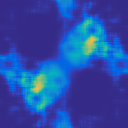}}};
        \end{scope}
    \end{tikzpicture}
\end{minipage}\hspace{.07em}
\begin{minipage}[c]{.235\textwidth}
\centering
\begin{tikzpicture}
        \node[anchor=south west,inner sep=0] (image) at (0,0) {\cfboxR{other3}{\includegraphics[width=.98\textwidth]{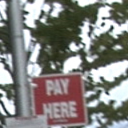}}};
        \begin{scope}[x={(image.south east)},y={(image.north west)}]
            \node[anchor=north east,inner sep=0] (image) at (1,1) {\cfboxR{other3}{\includegraphics[width=.35\textwidth]{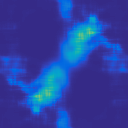}}};
        \end{scope}
    \end{tikzpicture}
\end{minipage}\hspace{.07em}
\begin{minipage}[c]{.235\textwidth}
\centering
\begin{tikzpicture}
        \node[anchor=south west,inner sep=0] (image) at (0,0) {\cfboxR{other4}{\includegraphics[width=.98\textwidth]{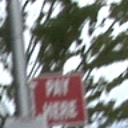}}};
        \begin{scope}[x={(image.south east)},y={(image.north west)}]
            \node[anchor=north east,inner sep=0] (image) at (1,1) {\cfboxR{other4}{\includegraphics[width=.35\textwidth]{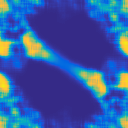}}};
        \end{scope}
    \end{tikzpicture}
\end{minipage}

\vspace{.2em}

\begin{minipage}[c]{.235\textwidth}
\centering
\begin{tikzpicture}
        \node[anchor=south west,inner sep=0] (image) at (0,0) {\cfboxR{other5}{\includegraphics[width=.98\textwidth]{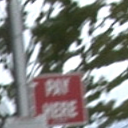}}};
        \begin{scope}[x={(image.south east)},y={(image.north west)}]
            \node[anchor=north east,inner sep=0] (image) at (1,1) {\cfboxR{other5}{\includegraphics[width=.35\textwidth]{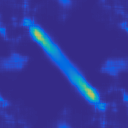}}};
        \end{scope}
    \end{tikzpicture}
\end{minipage}\hspace{.07em}
\begin{minipage}[c]{.235\textwidth}
\centering
\begin{tikzpicture}
        \node[anchor=south west,inner sep=0] (image) at (0,0) {\cfboxR{other6}{\includegraphics[width=.98\textwidth]{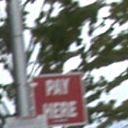}}};
        \begin{scope}[x={(image.south east)},y={(image.north west)}]
            \node[anchor=north east,inner sep=0] (image) at (1,1) {\cfboxR{other6}{\includegraphics[width=.35\textwidth]{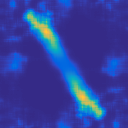}}};
        \end{scope}
    \end{tikzpicture}
\end{minipage}\hspace{.07em}
\begin{minipage}[c]{.235\textwidth}
\centering
\begin{tikzpicture}
        \node[anchor=south west,inner sep=0] (image) at (0,0) {\cfboxR{other7}{\includegraphics[width=.98\textwidth]{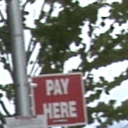}}};
        \begin{scope}[x={(image.south east)},y={(image.north west)}]
            \node[anchor=north east,inner sep=0] (image) at (1,1) {\cfboxR{other7}{\includegraphics[width=.35\textwidth]{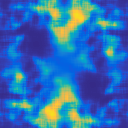}}};
        \end{scope}
    \end{tikzpicture}
\end{minipage}\hspace{.07em}
\begin{minipage}[c]{.235\textwidth}
\centering
\vspace{.6em}

{\ssmall  {\ssmall \hspace{.2em} Weights Distribution}} \vspace{.4em}

\includegraphics[width=.93\textwidth]{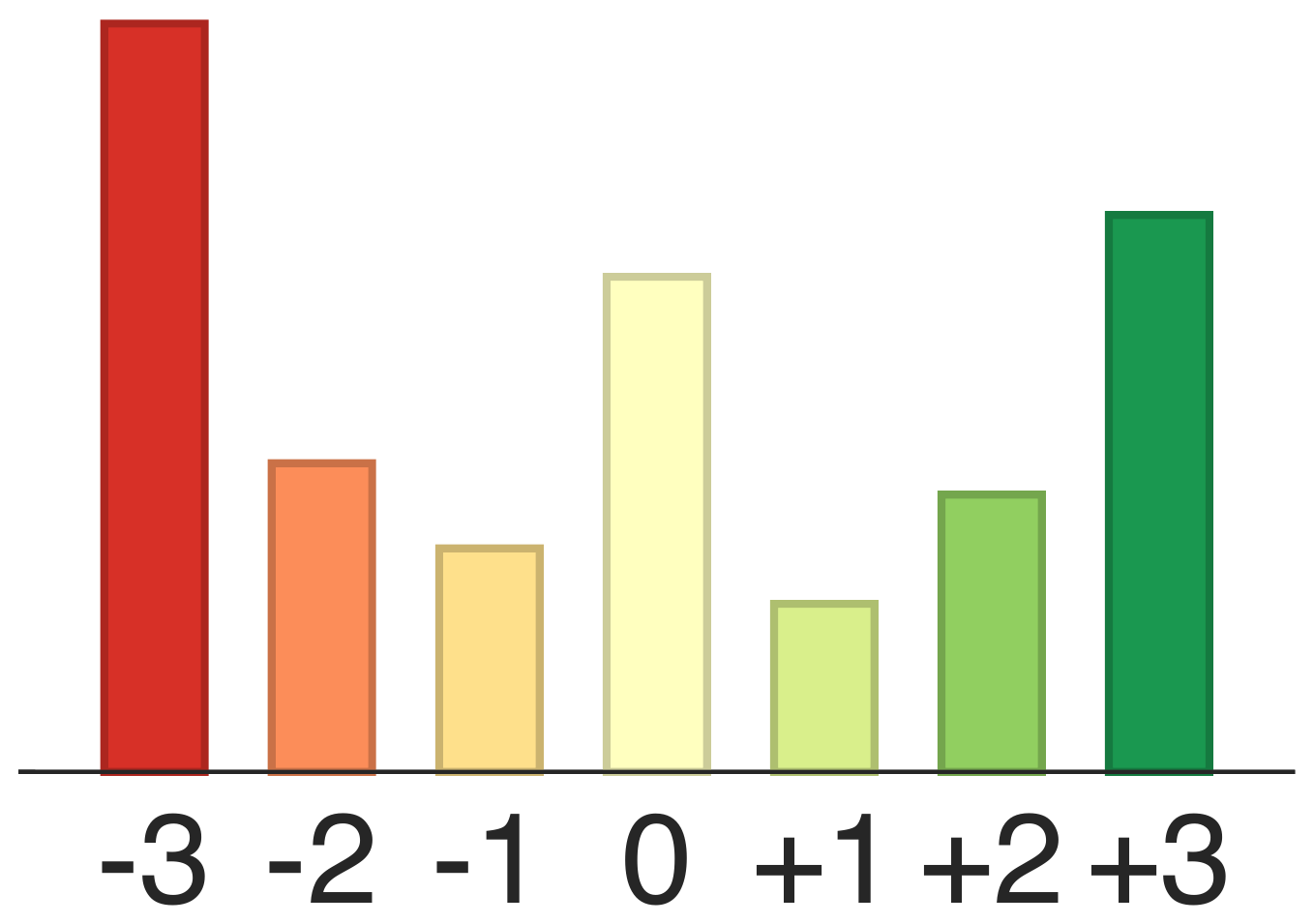}\vspace{0.2em}

\end{minipage}
\end{minipage}

\caption{An example of the intermediate results of the two main steps of the proposed algorithm. The top block shows an image crop from a (non registered) sequence of $7$ frames and the output of the algorithm for the reference frame. The bottom block shows the consistent registration of the image sequence with respect to the center frame (Ref). The output image results from the aggregation of different Fourier components present in different frames. This is confirmed by the Fourier weights distribution shown in the top-right corner of each frame. The bar plot on the bottom-right shows the frame contribution (by measuring the norm of the Fourier weights for each frame). }
\label{fig:runEx}
\end{figure}
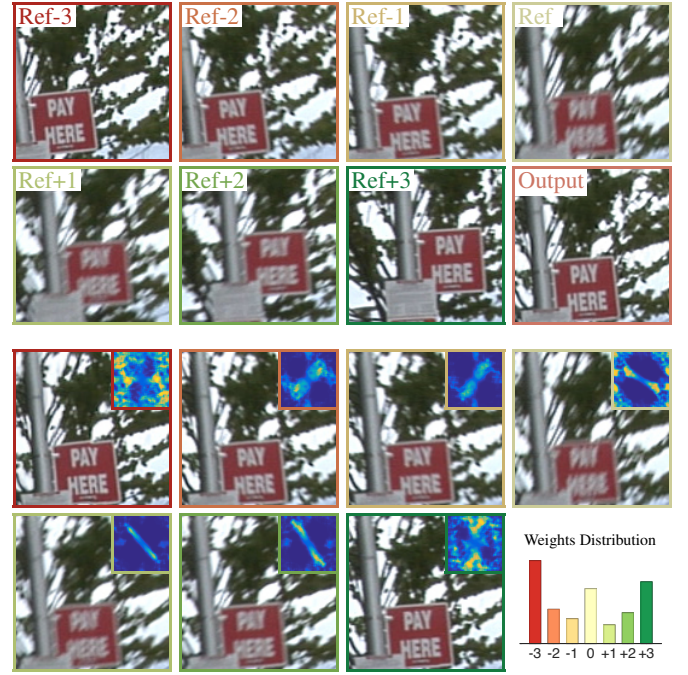

Figure~\ref{fig:runEx} shows an example of the intermediate results of the two main steps of the proposed algorithm. In this example, the output image results from the aggregation of different Fourier components present in different frames. This is confirmed by the Fourier weights distribution shown in the figure.

\subsection{Iterative Improvement}
Given a sequence of $N$ images $\{v_i\}_{i=1,\ldots,N}$, the previous two steps, produce a new sequence of $N$ images $\{\tilde{v}_i\}$. Each of these frames is created by combining the current frame and the $2M$ frames around it. In order to propagate the blur reduction to frames that are initially farther than $M$, we can proceed to apply the method iteratively.

The number of iterations needed depends on the sequence, and it is related to the type of blur, and how different the blur in nearby frames is. All the examples shown in this paper were computed with 1 to 4 iterations, but in most cases applying the method only once produces significantly better results over the input sequence.

\begin{figure}[tbp]
\ssmall
\centering

\begin{minipage}[c]{\columnwidth}
\begin{minipage}[c]{.187\columnwidth}
\centering
original%

\cfbox{red11}{\includegraphics[width=\textwidth]{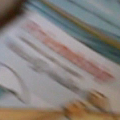}}
\end{minipage}
\hspace{.125em}
\begin{minipage}[c]{.187\columnwidth}
\centering
1st iteration\vspace{.2em}

\cfbox{red11}{\includegraphics[width=\textwidth]{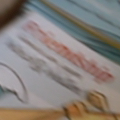}}
\end{minipage}
\hspace{.125em}
\begin{minipage}[c]{.187\columnwidth}
\centering
2nd iteration\vspace{.2em}

\cfbox{red11}{\includegraphics[width=\textwidth]{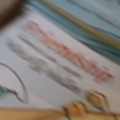}}
\end{minipage}
\hspace{.125em}
\begin{minipage}[c]{.187\columnwidth}
\centering
3rd iteration\vspace{.2em}

\cfbox{red11}{\includegraphics[width=\textwidth]{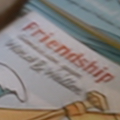}}
\end{minipage}
\hspace{.125em}
\begin{minipage}[c]{.187\columnwidth}
\centering
4th iteration\vspace{.2em}

\cfbox{red11}{\includegraphics[width=\textwidth]{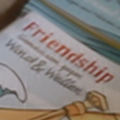}}
\end{minipage}

\vspace{.5em}
\end{minipage}
(a) 

\vspace{.5em}

\begin{minipage}[c]{0.75\columnwidth}
\centering
\includegraphics[width=\textwidth]{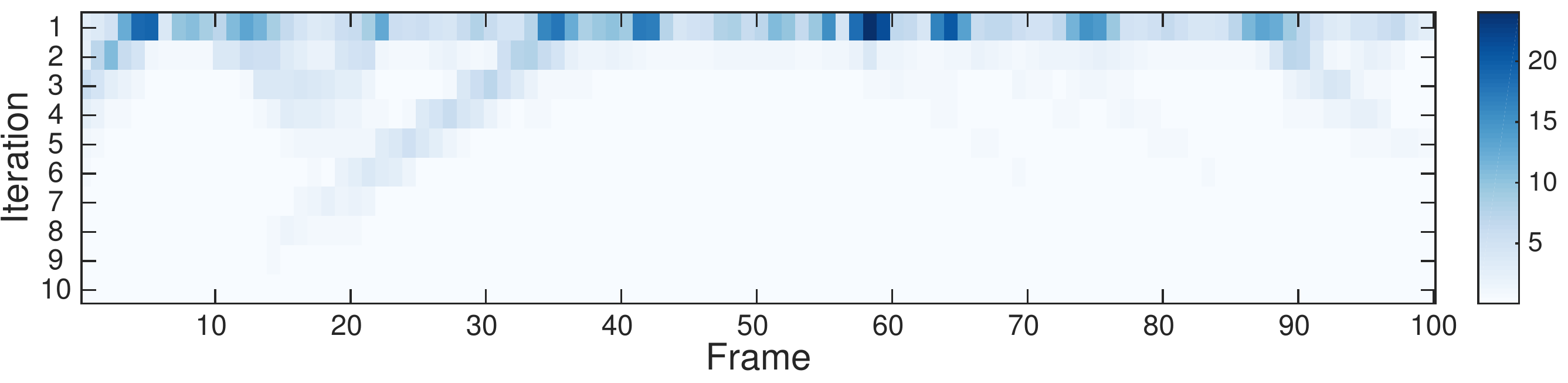}

(b)
\end{minipage}
\begin{minipage}[c]{0.23\columnwidth}
\centering
\includegraphics[width=\textwidth]{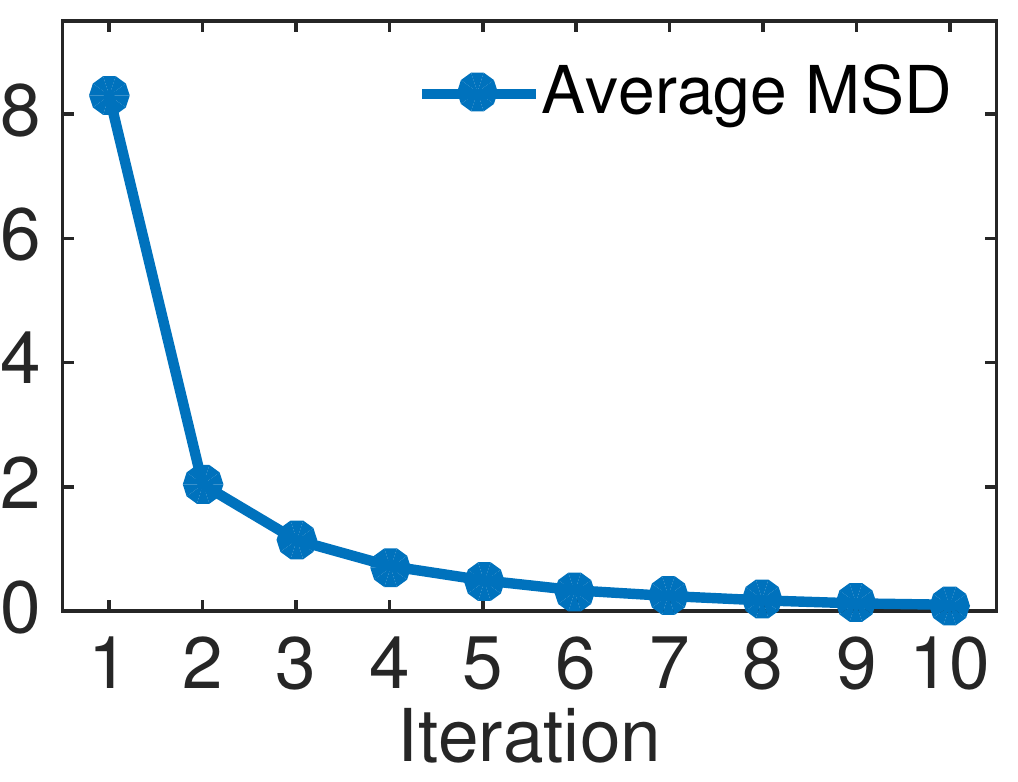}

(c)
\end{minipage}

\caption{Iterative restoration. (a) shows the result of iterating the proposed algorithm. This will propagate 
the information to frames that are initially farther than the given temporal window $[-M,M]$ frames. In this particular case four iterations are needed to significantly improve the quality of the frame. In (b) we show the square difference between a frame and the one from the previous iterations (first row shows the difference between the result of the first pass and the input sequence). Although most  frames do not change after the first iteration, a few change due to the update of the nearby frames. In particular, the plot shows some diagonal structure representing good frames that are transferring information to their nearby ones. (c) shows the average change of the whole sequence when iterating (each point is the average of each row in (b)). Most of the work is done in the first pass.  }
\label{fig:iterative}
\end{figure}

Figure~\ref{fig:iterative} shows an example of the effect of iteratively applying the deblurring algorithm. In this particular sequence, to get the best results in every frame four iterations are required. This is a very challenging sequence since most of the frames are significantly blurred and it has only a few very sparse sharp frames.
However, as shown in Figure~\ref{fig:iterative}b), most of the frames do not change a lot after the first pass. Nevertheless, there are some frames that continue to propagate information to nearby frames (see figure's caption for details).

Since the algorithm averages frames, and does not actually solve any inverse problem, at the end the video sequence may have some remaining blur. To enhance the final quality we can apply a simple unsharp masking step.

\begin{algorithm}[t]
\caption{Consistent Aggregation of a Sequence}
\label{algo:pAggregation}
\footnotesize
\Input{A sequence of $2M+1$ \textsc{rgb} images $v_{-\!M},\dotsc,v_0,\dotsc, v_M$ of size $m_h \times m_w \times n_c$, block size $b$, block overlap $s$, \textsc{fba} paramater $p$. }

\Output{Filtered (reference) image $\bar{u}$.}
\BlankLine

 \Comment{{\normalsize Consistent Registration}}                               
\For{$i\!=\!-M\!:\! M$} 
  {
 \BlankLine
  $\tau^0_i = \textsc{OpticalFlow}(v_i,v_0)$\Comment*{Forward flow estimation}
  $\tau^i_0 = \textsc{OpticalFlow}(v_0,v_i)$\Comment*{Backward flow estimation} \vspace{.4em}
  
  $\text{cMap}(\bx) =  |( \tau^0_i \circ \tau^i_0 )(\bx)   - \bx|$\Comment*{Consistent Pixels Map}
  
  ${M}(\bx) = \text{cMap}(\bx) \le \epsilon$\Comment*{Consistent Pixels Mask}
  ${M}(\bx) = G _\rho \left( \textsc{dilate} (M(\bx),r) \right)$\Comment*{Dilate and Smooth} \vspace{.5em}
  $v^0_i (\bx) = M(\bx)  \cdot ( v_i \circ \tau^0_i ) (\bx) + (1-M(\bx) ) \cdot v_0 (\bx)$\; 

\BlankLine
}
\vspace{.5em}

 \Comment{{\normalsize Local Fourier Burst Accumulation}}

 $u = \text{zeros}(m_h,m_w,n_c)$; 
 $c = \text{zeros}(m_h,m_w)$\Comment*{Aux. Buffers Initialization}
 \vspace{.5em}
 
  \For{ $j\!=\!1\!:\!s\!:\!m_h$ \normalfont{\textbf{ and }} $k\!=\!1\!:\!s\!:\!m_w$} 
 {
   $\hat{Q} = \text{zeros}(b,b,n_c)$;
   $w = \text{zeros}(b,b)$\Comment*{Aux. Block Buffer Initialization}\vspace{.2em}
   $X_{j,k} =   \text{coordinates of } (b\!\times \! b\!\times\! n_c$)-patch centered at pixel $(k,l)$\;
   
\For{$i\!=\!-M\!:\! M$} 
   {
        $P_i = v^0_i(X_{j,k})$\;

       $\hat{P}_i = \textsc{fft}(P_i)$\;
       $w_i = \text{colorAverage}\left( |\hat{P}_i | \right)$\Comment*{Mean over color channels}
       $w_i = G_\sigma w_i$ \Comment*{Gaussian smoothing}

       $\hat{Q} =  \hat{Q}+ w^p_i \cdot \hat{P}_i$\Comment*{Weighted Block Fourier Accumulation}
       
       $w = w + w^p_i$\;
   }
   
    $Q =  \textsc{ifft} ( \hat{Q} ./ w)$\Comment*{Estimation of pixel values of block $X_{k,j}$}
   
    $u(X_{j,k}) = u(X_{j,k}) + Q$\;
    $c(X_{j,k}) = c(X_{j,k}) + 1$\;

}

$u =  u./c$\;

\vspace{.5em}

\Comment{{\footnotesize \textbf{Comments:}  $u(X_{j,k})$ is the evaluation of $u$ on each pixel in patch $X_{j,k}$. The operator $./$ (lines 19 and 22) represents element-wise division. The notation $j=1\!:\!s\!:\!m$ implies that $j$ takes the integer values from 1 to $m$ by increments of $s$. $G_\sigma$ represents a Gaussian Smoothing of standard deviation $\sigma$. In the current implementation,  $\sigma =  \nicefrac{50}{b}$ and $\rho=5$. The dilatation operation (line 6) is done with a circular element of radius $r=5$. Image warpings are done via bicubic interpolation. The consistent registration tolerance is set to $\epsilon=1$.
Default values: $M=3$, $b=128$, $s=64$, and $p=11$.}}
\end{algorithm}

\subsection{Complexity Analysis and Execution Time}
Let  $m = m_h \times m_w$ be the number of image pixels, $B = b \times b$ the block size, and $2M+1$ the number of consecutive frames use in the temporal window. If we operate with half-overlapped blocks ($s=b/2$ in Algorithm 1) then, the more demanding part is the computation of all the Fourier Transforms, namely $O((2M+1) \cdot 2 m \cdot \log B)$. This is the reason to the very low complexity of the method. In addition, one has to compute the Gaussian smoothing of the weights and the power to the $p$ that are linear operators on the number of image pixels. 

Regarding memory consumption, the algorithm does not need to access all the images simultaneously, and can proceed in an online fashion. However, for simplicity, we keep all the registered sequence in memory to speed up the access to the image blocks. In addition, four buffers are needed: two of the size of a video frame and two  of the size of the block (see Algorithm~\ref{algo:pAggregation}). 

Our Matlab prototype takes about 15 seconds to filter a HD frame on a MacBook Pro 2.6Ghz i5. This is with the default parameters: $M=3$ (7 frames), block size $b_h=b_w=128$, and blocks half-overlap. Two-thirds of the processing time are due to the optical flow computation and the consistent registration.
Regarding the filtering part, it can be highly accelerated, since the three key components (\textsc{fft}, Gaussian filtering, and power to the $p$) can be easily implemented in GPU.

\section{Experimental Results}
\label{sec:results}
To evaluate and compare the proposed method we used the seven video sequences provided by Cho et al.~\cite{cho2012video} as a basis, 
also showing additional results on a set of eight videos that were captured by us. These sequences show different amount of camera shake blur in varying circumstances: outdoors/indoors scenes, static scenes, moving objects, object occlusions. The full processed sequences and a video showing the results are available at the project's website.\footnote{\url{http://dev.ipol.im/~mdelbra/videoFA/}}
All the results were computed using the default parameters shown in Algorithm 1.

\vspace{.5em}

\begin{figure*}
\ssmall

\begin{minipage}[c]{\textwidth}
\centering

\begin{minipage}[c]{.255\textwidth}
\centering
\includegraphics[width=\textwidth]{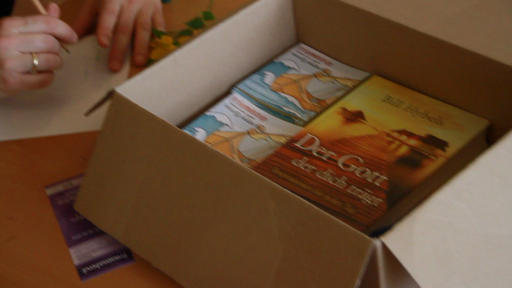} \vspace{-.9em}

\includegraphics[width=\textwidth]{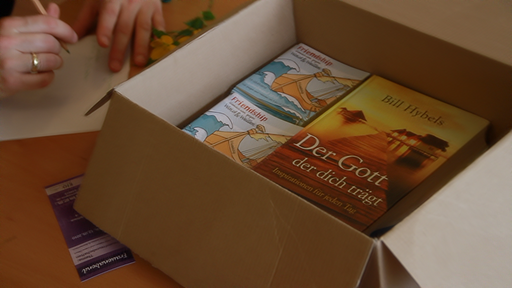}

\end{minipage}
\begin{minipage}[c]{.144\textwidth}
\centering

\includegraphics[width=\textwidth]{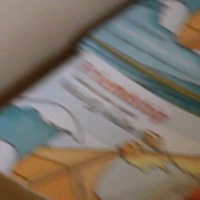} \vspace{-.9em}

\includegraphics[width=\textwidth]{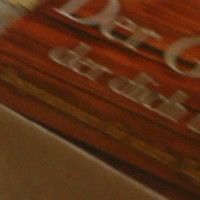}

\end{minipage}
%
%
%
%
\begin{minipage}[c]{.144\textwidth}
\centering

\includegraphics[width=\textwidth]{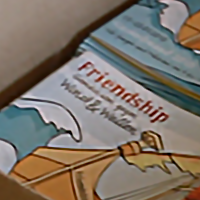} \vspace{-.9em}

\includegraphics[width=\textwidth]{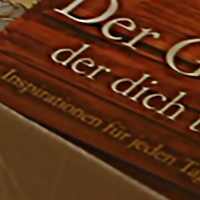}

\end{minipage}
\begin{minipage}[c]{.144\textwidth}
\centering

\includegraphics[width=\textwidth]{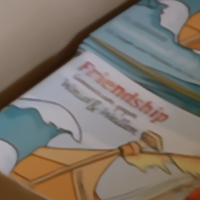} \vspace{-.9em}

\includegraphics[width=\textwidth]{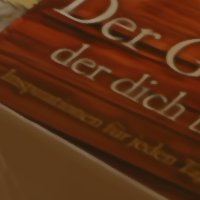}

\end{minipage}
\begin{minipage}[c]{.144\textwidth}
\centering

\includegraphics[width=\textwidth]{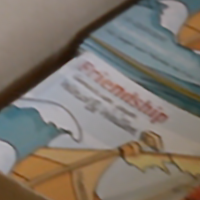} \vspace{-.9em}

\includegraphics[width=\textwidth]{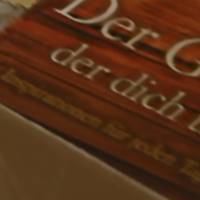}

\end{minipage}
\begin{minipage}[c]{.144\textwidth}
\centering

\includegraphics[width=\textwidth]{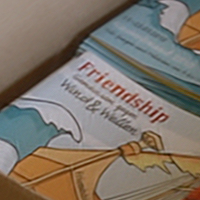} \vspace{-.9em}

\includegraphics[width=\textwidth]{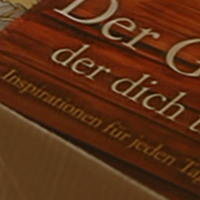}

\end{minipage}

\end{minipage}

\vspace{.5em}


\begin{minipage}[c]{\textwidth}
\centering

\begin{minipage}[c]{.255\textwidth}
\centering
\includegraphics[width=\textwidth]{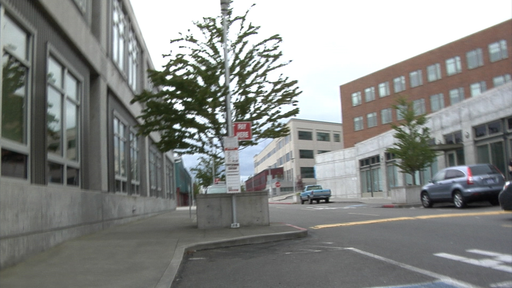} \vspace{-.9em}

\includegraphics[width=\textwidth]{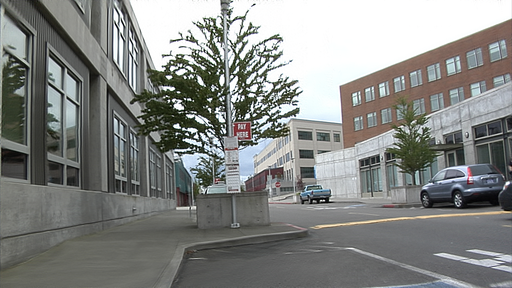}

\end{minipage}
\begin{minipage}[c]{.144\textwidth}
\centering

\includegraphics[width=\textwidth]{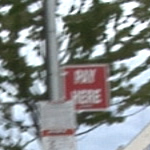} \vspace{-.9em}

\includegraphics[width=\textwidth]{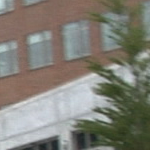}

\end{minipage}
%
%
%
%
\begin{minipage}[c]{.144\textwidth}
\centering

\includegraphics[width=\textwidth]{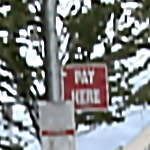} \vspace{-.9em}

\includegraphics[width=\textwidth]{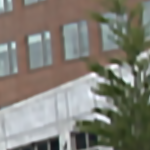}

\end{minipage}
\begin{minipage}[c]{.144\textwidth}
\centering

\includegraphics[width=\textwidth]{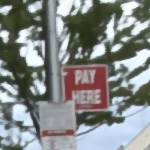} \vspace{-.9em}

\includegraphics[width=\textwidth]{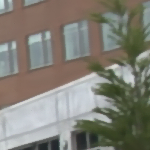}

\end{minipage}
\begin{minipage}[c]{.144\textwidth}
\centering

\includegraphics[width=\textwidth]{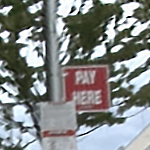} \vspace{-.9em}

\includegraphics[width=\textwidth]{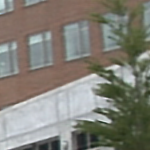}

\end{minipage}
\begin{minipage}[c]{.144\textwidth}
\centering

\includegraphics[width=\textwidth]{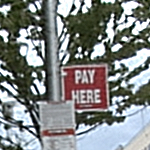} \vspace{-.9em}

\includegraphics[width=\textwidth]{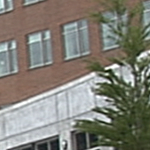}

\end{minipage}

\end{minipage}

\vspace{.5em}

\begin{minipage}[c]{\textwidth}
\centering

\begin{minipage}[c]{.255\textwidth}
\centering
\includegraphics[width=\textwidth]{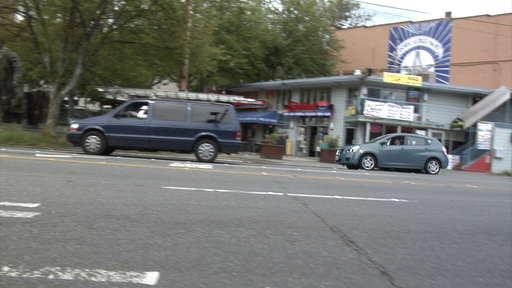} \vspace{-.9em}

\includegraphics[width=\textwidth]{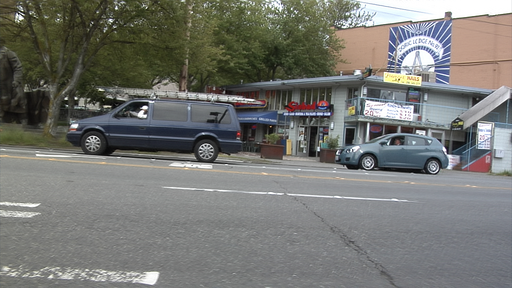}

\end{minipage}
\begin{minipage}[c]{.144\textwidth}
\centering

\includegraphics[width=\textwidth]{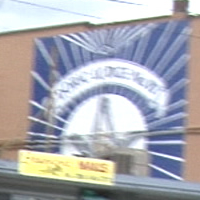} \vspace{-.9em}

\includegraphics[width=\textwidth]{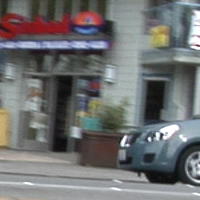}

\end{minipage}
%
%
%
\begin{minipage}[c]{.144\textwidth}
\centering

\includegraphics[width=\textwidth]{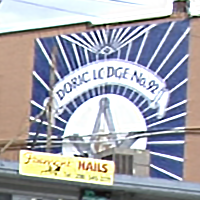} \vspace{-.9em}

\includegraphics[width=\textwidth]{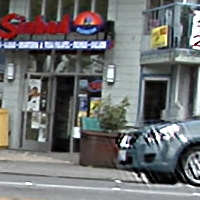}

\end{minipage}
\begin{minipage}[c]{.144\textwidth}
\centering

\includegraphics[width=\textwidth]{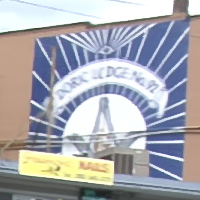} \vspace{-.9em}

\includegraphics[width=\textwidth]{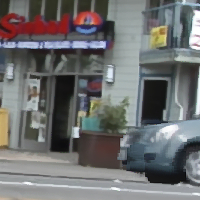}

\end{minipage}
\begin{minipage}[c]{.144\textwidth}
\centering

\includegraphics[width=\textwidth]{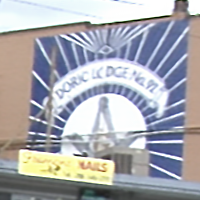} \vspace{-.9em}

\includegraphics[width=\textwidth]{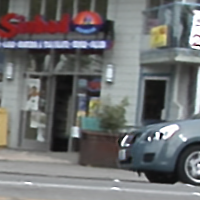}

\end{minipage}
\begin{minipage}[c]{.144\textwidth}
\centering

\includegraphics[width=\textwidth]{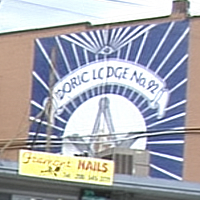} \vspace{-.9em}

\includegraphics[width=\textwidth]{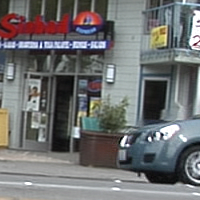}

\end{minipage}

\end{minipage}

\vspace{.5em}

\begin{minipage}[c]{\textwidth}
\centering

\begin{minipage}[c]{.255\textwidth}
\centering
\includegraphics[width=\textwidth]{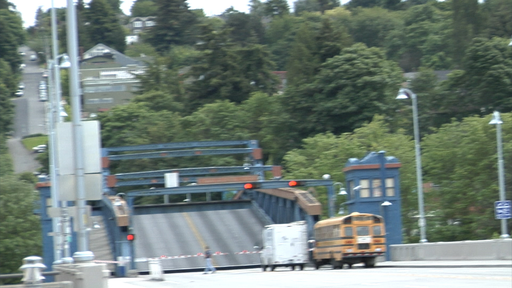} \vspace{-.9em}

\includegraphics[width=\textwidth]{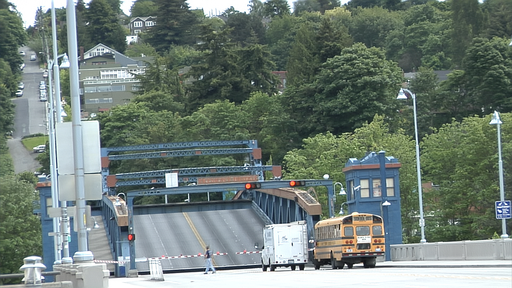}

{\scriptsize Blurry  (top) and \textbf{processed} (bottom) frames}

\end{minipage}
\begin{minipage}[c]{.144\textwidth}
\centering

\includegraphics[width=\textwidth]{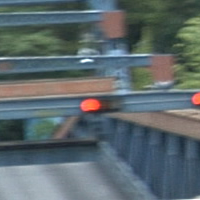} \vspace{-.9em}

\includegraphics[width=\textwidth]{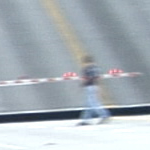}

{\scriptsize Blurry crop}

\end{minipage}
%
%
%
%
%
\begin{minipage}[c]{.144\textwidth}
\centering

\includegraphics[width=\textwidth]{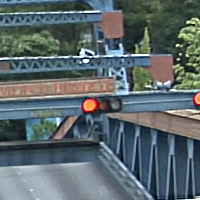} \vspace{-.9em}

\includegraphics[width=\textwidth]{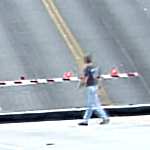}

{\scriptsize Zhang et al.~\cite{zhang2013multi}}

\end{minipage}
\begin{minipage}[c]{.144\textwidth}
\centering

\includegraphics[width=\textwidth]{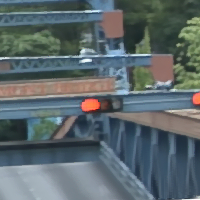} \vspace{-.9em}

\includegraphics[width=\textwidth]{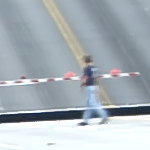}

{\scriptsize Cho et al.~\cite{cho2012video}}

\end{minipage}
\begin{minipage}[c]{.144\textwidth}
\centering

\includegraphics[width=\textwidth]{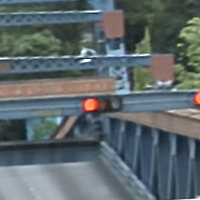} \vspace{-.9em}

\includegraphics[width=\textwidth]{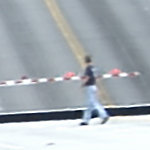}

{\scriptsize Kim and Lee~\cite{kim2015cvpr}}

\end{minipage}
\begin{minipage}[c]{.144\textwidth}
\centering

\includegraphics[width=\textwidth]{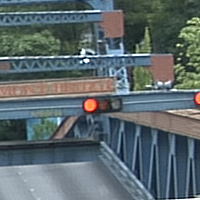} \vspace{-.9em}

\includegraphics[width=\textwidth]{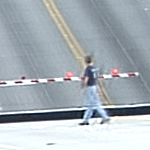}

{\scriptsize \textbf{Proposed method}}

\end{minipage}
\end{minipage}

\caption{Comparison to other deblurring methods I (rows 1-2 {\footnotesize \textsf{books} }seq., rows 3-4 {\footnotesize \textsf{street}} seq., rows 5-6 {\footnotesize\textsf{car}} seq., rows 7-8 {\footnotesize\textsf{bridge}} seq.) }
\label{fig:comparison1}
 
\end{figure*}


\begin{figure*}
\ssmall

\begin{minipage}[c]{\textwidth}
\centering

\begin{minipage}[c]{.255\textwidth}
\centering
\includegraphics[width=\textwidth]{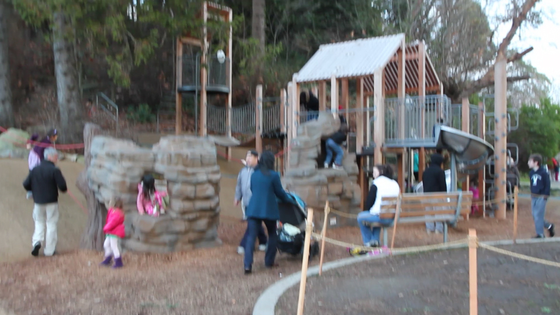} \vspace{-.9em}

\includegraphics[width=\textwidth]{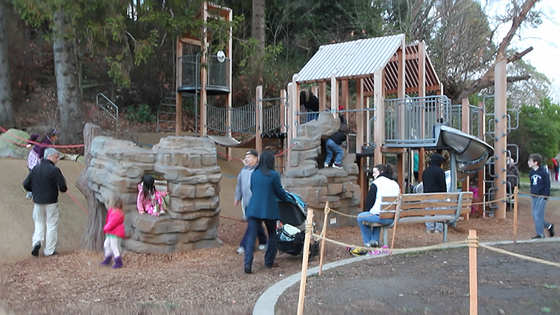}

\end{minipage}
\begin{minipage}[c]{.144\textwidth}
\centering

\includegraphics[width=\textwidth]{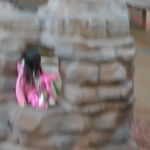} \vspace{-.9em}

\includegraphics[width=\textwidth]{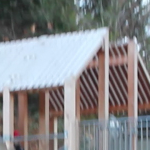}

\end{minipage}
\begin{minipage}[c]{.144\textwidth}
\centering

\includegraphics[width=\textwidth]{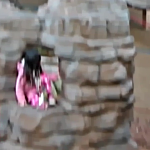} \vspace{-.9em}

\includegraphics[width=\textwidth]{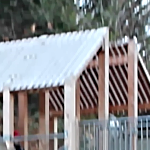}

\end{minipage}
\begin{minipage}[c]{.144\textwidth}
\centering

\includegraphics[width=\textwidth]{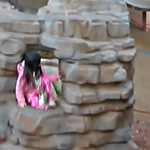} \vspace{-.9em}

\includegraphics[width=\textwidth]{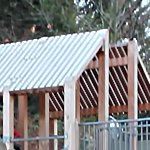}

\end{minipage}
\begin{minipage}[c]{.144\textwidth}
\centering

\includegraphics[width=\textwidth]{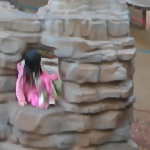} \vspace{-.9em}

\includegraphics[width=\textwidth]{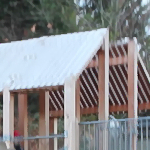}

\end{minipage}
\begin{minipage}[c]{.144\textwidth}
\centering

\includegraphics[width=\textwidth]{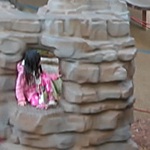} \vspace{-.9em}

\includegraphics[width=\textwidth]{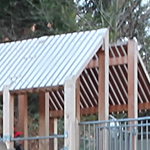}

\end{minipage}

\end{minipage}

\vspace{.5em}

\begin{minipage}[c]{\textwidth}
\centering

\begin{minipage}[c]{.255\textwidth}
\centering
\includegraphics[width=\textwidth, clip=true, trim=0 6 0 6]{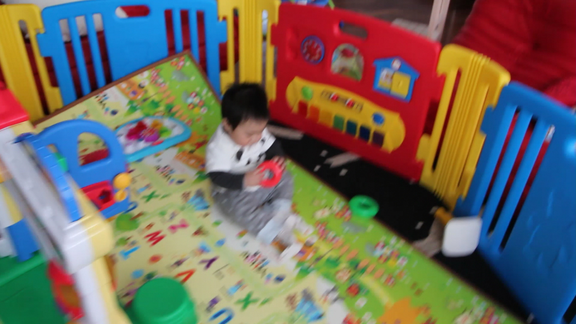}  \vspace{-.9em}

\includegraphics[width=\textwidth, clip=true, trim=0 6 0 6]{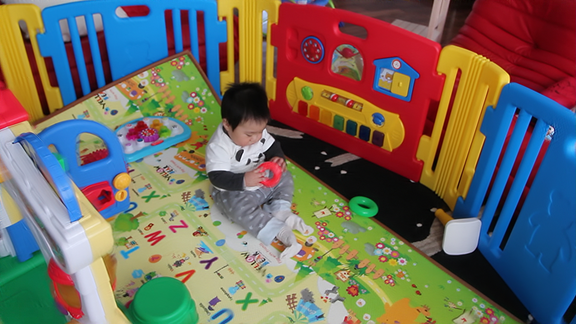}

{\scriptsize Blurry (top) and \textbf{processed} (bottom) frames}

\end{minipage}
\begin{minipage}[c]{.144\textwidth}
\centering

\includegraphics[width=\textwidth]{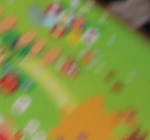} \vspace{-.9em}

\includegraphics[width=\textwidth]{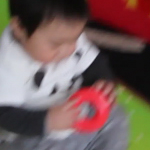}

{\scriptsize Blurry crop}

\end{minipage}
\begin{minipage}[c]{.144\textwidth}
\centering

\includegraphics[width=\textwidth]{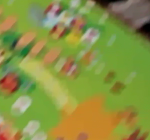} \vspace{-.9em}

\includegraphics[width=\textwidth]{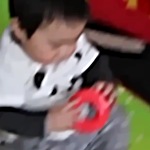}

{\scriptsize Krishnan et al.~\cite{krishnan2011blind}}

\end{minipage}
\begin{minipage}[c]{.144\textwidth}
\centering

\includegraphics[width=\textwidth]{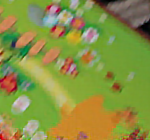} \vspace{-.9em}

\includegraphics[width=\textwidth]{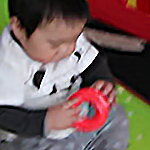}

{\scriptsize Zhang et al.~\cite{zhang2013multi}}

\end{minipage}
\begin{minipage}[c]{.144\textwidth}
\centering

\includegraphics[width=\textwidth]{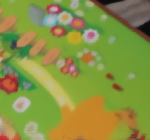} \vspace{-.9em}

\includegraphics[width=\textwidth]{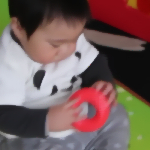}

{\scriptsize Cho et al.~\cite{cho2012video}}

\end{minipage}
\begin{minipage}[c]{.144\textwidth}
\centering

\includegraphics[width=\textwidth]{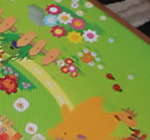} \vspace{-.9em}

\includegraphics[width=\textwidth]{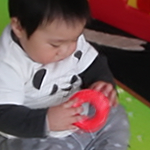}

{\scriptsize \textbf{Proposed method}}

\end{minipage}

\end{minipage}

\caption{Comparison to other deblurring methods II (rows 1-2 {\footnotesize \textsf{playground} }seq., rows 3-4 {\footnotesize \textsf{kids}} seq.) }
\label{fig:comparison2}
 
\end{figure*}

\noindent \textbf{Comparison to other video deblurring methods.}
We compared the proposed algorithm to four other methods, both regarding image quality and execution time. The first one is the single image deconvolution algorithm by Krishnan et al.~\cite{krishnan2011blind}.  This algorithm introduces, as a natural image prior, the ratio between the $\ell_1$ and the $\ell_2$ norms on the high frequencies of an image. This normalized sparsity measure gives low cost for the sharp image. 
Second, we compare to the multi-image deconvolution algorithm by Zhang et al.~\cite{zhang2013multi}. This algorithm proposes a Bayesian framework for coupling all the unknown blurring kernels and the latent sharp image in a single prior. To avoid introducing image artifacts due to moving objects and occlusions (which their algorithm is not designed to handle), we run this algorithm on the set of consistent registered frames. Although this formulation in general produces good looking sharp images, its optimization is very slow and requires several minutes for filtering an HD frame using 7 nearby frames. For both deconvolution algorithms we used the code provided by the authors. The algorithms rely on parameters that were manually tuned to get the best possible results.
Third, we compare our results to the video deblurring method by Cho et al.~\cite{cho2012video}. This method is conceptually similar to ours, since it proposes to transfer information from nearby frames to restore the quality of each frame in the video. The results are the ones provided by the authors.
Finally, we compare to the very recent algorithm by Kim and Lee~\cite{kim2015cvpr}. This method jointly estimates the optical flow and latent sharp frames by minimizing an energy function penalizing inconsistencies to a forward model. The method is general in the sense that is (potentially) capable of 
removing any blur given by the estimation of the optical flow and the camera duty cycle. The adopted energy function has several regularization terms that forces
spatial and temporal consistency. The results are the ones provided by the authors.

Figures \ref{fig:comparison1} and \ref{fig:comparison2} show some selected crops for 6 different restored videos (provided by Cho et al.~\cite{cho2012video}).  These figures show that the proposed method can successfully remove camera shake blur in realistic scenarios. In general, the proposed algorithm obtains similar or better results than those from the multi-image deconvolution algorithm by Zhang et al.~\cite{zhang2013multi}, at significantly reduced computational cost. Although~\cite{zhang2013multi} produces sharp images, it sometimes creates artifacts. This is a result of trying to solve an inverse problem with an inaccurately estimated forward model (e.g., the blurring kernels). This is clearly observed in the ``pay here'' sign (Figure 6, third row) or in the kid's carpet (Figure 7, third row). In addition, due to the  required complex optimization, this algorithm takes several minutes to filter a single frame virtually prohibiting its use for restoring full video sequences.

The single image deconvolution method in~\cite{krishnan2011blind} manages to get sharper images than the input ones, but their quality is significantly lower to the ones produced by our method. The main reason is that this algorithm does not use any information from the nearby --possibly sharp-- frames.

The video deblurring algorithm proposed by Cho et al. manages to get good clean results. However, similar to other non-local based restoration methods, the results  are often oversmooth due to the averaging of many different patches. Indeed, the extension to deal with video blur is very challenging, since the algorithm needs to find patches that are similar but differently blurred. This is observed, for example, in the books sequence (Figure 6, first/second rows) where it is impossible to read most of the text. In addition, the proposed algorithm is much faster since it does not require to compare patches, a highly computationally demanding task.

The general video deblurring algorithm proposed by Kim and Lee~\cite{kim2015cvpr} produces in general good quality results. Nevertheless, due to the strong imposed regularization, in many situations, the results  present  cartoon  artifacts due to the  conventional total variation image prior (to successfully remove blur total variation regularization tends to generate regions of constant color, separated by edges). This is observed, for example, in the {\small \textsf{streets}} and {\small \textsf{car}} sequences (Figure 6) where many details have been flattened. Additionally, due to the complex non-convex minimization, this algorithm requires significant computation power taking approximately 12 minutes to process a single HD frame.

\vspace{.3em}

\begin{figure}[tpb]
\centering

\begin{minipage}[c]{.243\columnwidth}

\includegraphics[width=\textwidth]{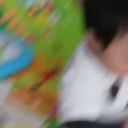} \vspace{-.93em}

\includegraphics[width=\textwidth]{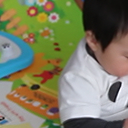}

\end{minipage}\hspace{-.1em}
\begin{minipage}[c]{.243\columnwidth}

\includegraphics[width=\textwidth]{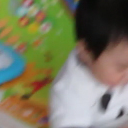} \vspace{-.93em}

\includegraphics[width=\textwidth]{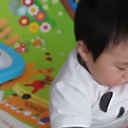}

\end{minipage}\hspace{-.1em}
\begin{minipage}[c]{.243\columnwidth}

\includegraphics[width=\textwidth]{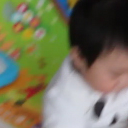} \vspace{-.93em}

\includegraphics[width=\textwidth]{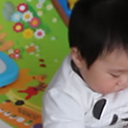}

\end{minipage}\hspace{-.1em}
\begin{minipage}[c]{.243\columnwidth}

\includegraphics[width=\textwidth]{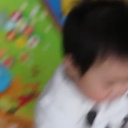} \vspace{-.93em}

\includegraphics[width=\textwidth]{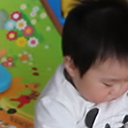}

\end{minipage}

\vspace{.6em}

\begin{minipage}[c]{.243\columnwidth}

\includegraphics[width=\textwidth]{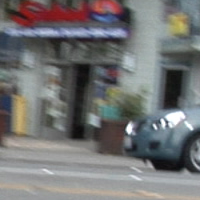} \vspace{-.93em}

\includegraphics[width=\textwidth]{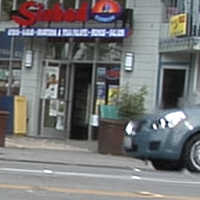}

\end{minipage}\hspace{-.1em}
\begin{minipage}[c]{.243\columnwidth}

\includegraphics[width=\textwidth]{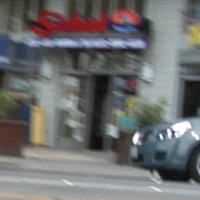} \vspace{-.93em}

\includegraphics[width=\textwidth]{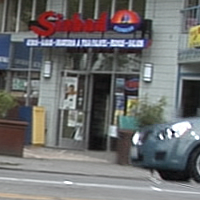}

\end{minipage}\hspace{-.1em}
\begin{minipage}[c]{.243\columnwidth}

\includegraphics[width=\textwidth]{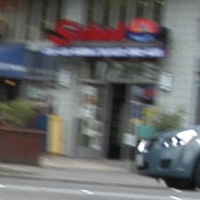}  \vspace{-.93em}

\includegraphics[width=\textwidth]{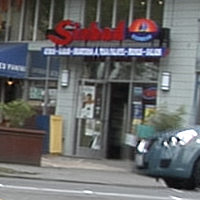}

\end{minipage}\hspace{-.1em}
\begin{minipage}[c]{.243\columnwidth}

\includegraphics[width=\textwidth]{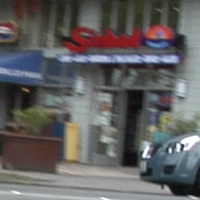}  \vspace{-.93em}

\includegraphics[width=\textwidth]{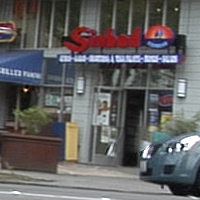}

\end{minipage}

\caption{Examples of consecutive filtered frames from two sequences ({\scriptsize \textsf{car}} and {\scriptsize \textsf{kid}}). The top row shows image crops of the input blurry sequence, while the bottom row the proposed algorithm's results.}
\label{fig:seqEx}

\end{figure}

\begin{figure}[tpb]
\footnotesize
\centering

\begin{minipage}[c]{.01\columnwidth}

\begin{sideways}
\hspace{1.5em}
\textbf{with CR}
\hspace{3.5em}
without CR
\hspace{3em}
Input
\hspace{2em}
\end{sideways}
\end{minipage}
\hspace{.2em}
\begin{minipage}[c]{.23\columnwidth}

\cfboxR{other1}{\includegraphics[width=.98\textwidth]{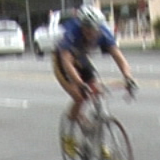}} \vspace{-.5em}

\cfboxR{other3c}{\includegraphics[width=.98\textwidth]{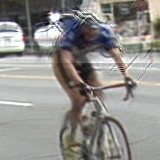}}  \vspace{-.5em}

\cfboxR{other7}{\includegraphics[width=.98\textwidth]{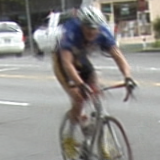}}

\end{minipage}
\begin{minipage}[c]{.23\columnwidth}

\cfboxR{other1}{\includegraphics[width=.98\textwidth]{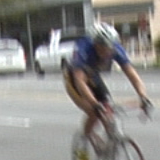}}  \vspace{-.5em}

\cfboxR{other3c}{\includegraphics[width=.98\textwidth]{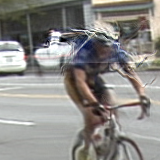}}  \vspace{-.5em}

\cfboxR{other7}{\includegraphics[width=.98\textwidth]{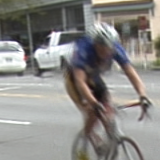}}

\end{minipage}
\begin{minipage}[c]{.23\columnwidth}

\cfboxR{other1}{\includegraphics[width=.98\textwidth]{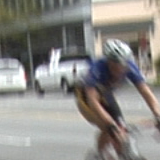}}  \vspace{-.5em}

\cfboxR{other3c}{\includegraphics[width=.98\textwidth]{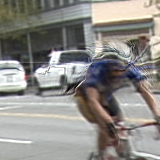}}  \vspace{-.5em}

\cfboxR{other7}{\includegraphics[width=.98\textwidth]{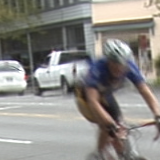}}

\end{minipage}
\begin{minipage}[c]{.23\columnwidth}

\cfboxR{other1}{\includegraphics[width=.98\textwidth]{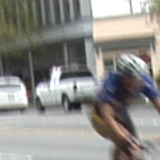}}  \vspace{-.5em}

\cfboxR{other3c}{\includegraphics[width=.98\textwidth]{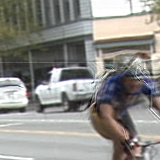}}  \vspace{-.5em}

\cfboxR{other7}{\includegraphics[width=.98\textwidth]{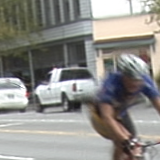}}

\end{minipage}

\caption{The importance of the consistent registration (CR). If the video sequence is registered directly using an optical flow estimation that does not consider occlusions or moving objects, the frame fusion will have artifacts (second row: \emph{without CR}). This is avoided by the proposed consistent registration that detect pixels not having symmetric optical flow estimations (third row: \emph{with CR}). The filtered sequence does not have artifacts. Instead it keeps the moving object unaltered. See supplementary 
material to observe the sharp quality of the processed movie while at the same time maintaining 
spatial and temporal coherence.}
\label{fig:seqNoCon}
\end{figure}

\noindent \textbf{Consistent registration and temporal coherence.}
In figures~\ref{fig:seqEx} and \ref{fig:seqNoCon} we show several frame crops of two of the considered video sequences, for the input blurry sequences and the proposed algorithm's results. As a general observation, the output images are much sharper than the input ones. The bike sequence shown in Figure \ref{fig:seqNoCon} is particularly challenging due to the biker's movement and the cars in the background. In this sequence, we can see the importance of the consistent registration to avoid creating image artifacts.

Note that while we do not explicitly force any temporal coherence, the filtered sequences are in general temporally coherent. Since the Fourier weighting scheme is done in a moving temporal window, the filtering yields results that are naturally temporally coherent. This can be checked in the videos provided in the supplementary material.

\vspace{.3em}

\begin{figure}[tpb]
\footnotesize
\centering

\begin{minipage}[c]{\columnwidth}
\centering
\begin{minipage}[c]{.28\columnwidth}
\centering
Input

\cfbox{red11}{\includegraphics[width=\textwidth]{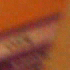}}
\end{minipage}
\hspace{.125em}
\begin{minipage}[c]{.28\columnwidth}
\centering
1st iteration\vspace{.2em}

\cfbox{red11}{\includegraphics[width=\textwidth]{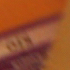}}
\end{minipage}
\hspace{.125em}
\begin{minipage}[c]{.28\columnwidth}
\centering
2nd iteration\vspace{.2em}

\cfbox{red11}{\includegraphics[width=\textwidth]{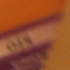}}
\end{minipage}

\vspace{.5em}
\end{minipage}
(a) 

\vspace{.8em}

\begin{minipage}[c]{0.75\columnwidth}
\centering
\includegraphics[width=\textwidth]{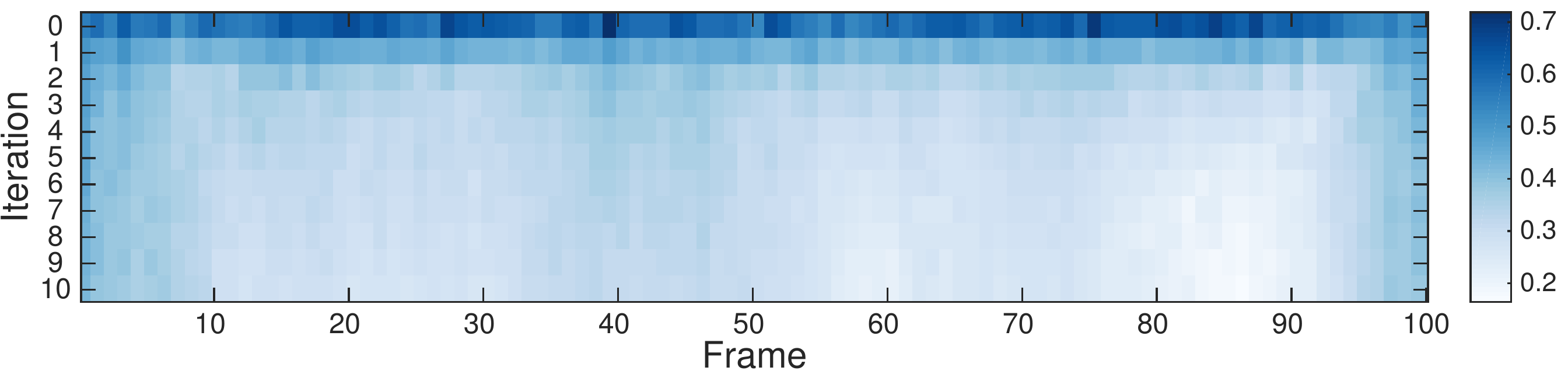}

(b)
\end{minipage}
\begin{minipage}[c]{0.23\columnwidth}
\centering
\includegraphics[width=\textwidth]{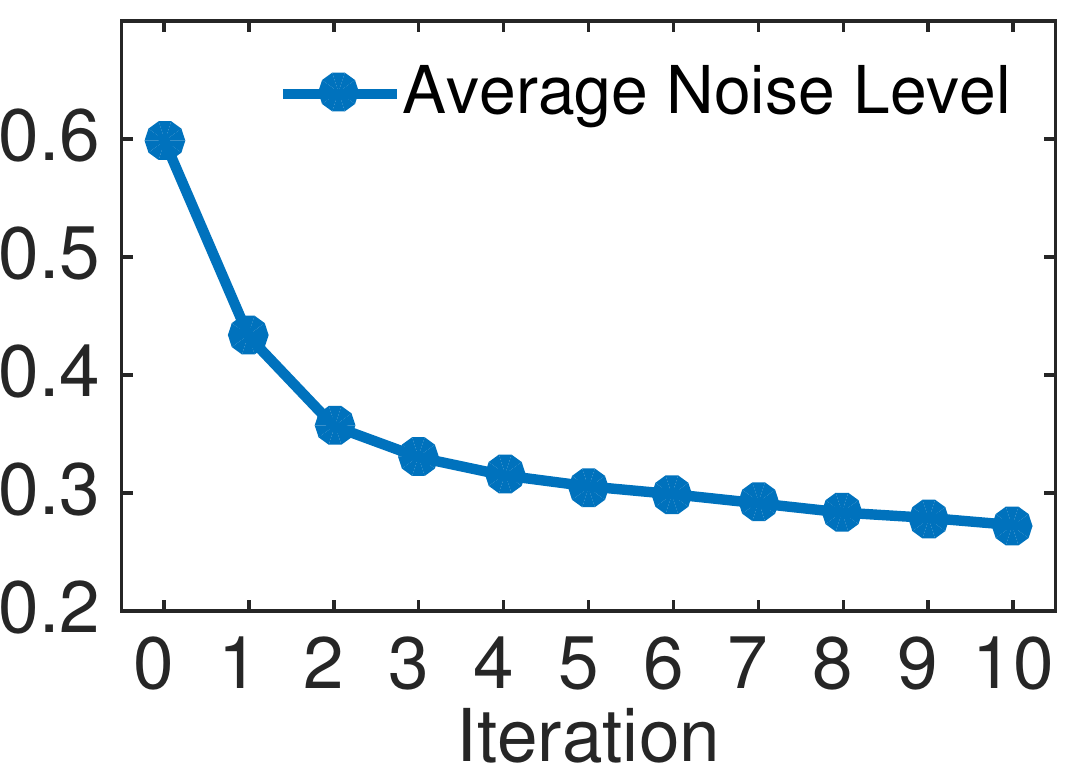}

(c)
\end{minipage}

\caption{Noise reduction as a byproduct effect.  When different frames are fusioned, different realizations of noise are
averaged leading to a noise reduction. In the example shown in (a), after the first pass of the algorithm the noise is significantly reduced. In (b), we show an estimation of the image noise level at different iterations for every frame in the sequence. 
The estimation is done using~\protect\cite{colom2013analysis}.
(c) shows the average noise level in the whole sequence at each  iteration (each point is the average of each row in (b)). The first pass is the one having a larger denoising effect since it is averaging completely independent realizations. 
}
\label{fig:iterativeNoise}
\end{figure}

\begin{figure}[tpb]
\footnotesize
\centering
\begin{minipage}[c]{.49\columnwidth}
\centering

Input (noisy) frame
\vspace{.1em}

\begin{tikzpicture}
    \node[anchor=north west,inner sep=0] (image) at (0,0) {\includegraphics[width=\textwidth]{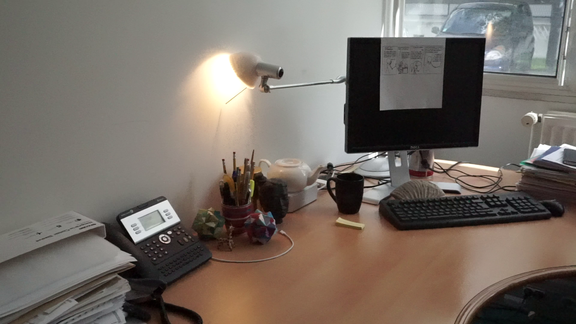}};
    \begin{scope}[x={(image.north east)},y={(image.south west)}]
        \draw[other1b, very thick] (0.35,0.6) rectangle (0.45,0.8); 
    \end{scope}
\end{tikzpicture}
\end{minipage}
\hspace{-.5em}
\begin{minipage}[c]{.49\columnwidth}
\centering

Processed frame
\vspace{.3em}

\begin{tikzpicture}
    \node[anchor=north west,inner sep=0] (image) at (0,0) {\includegraphics[width=\textwidth]{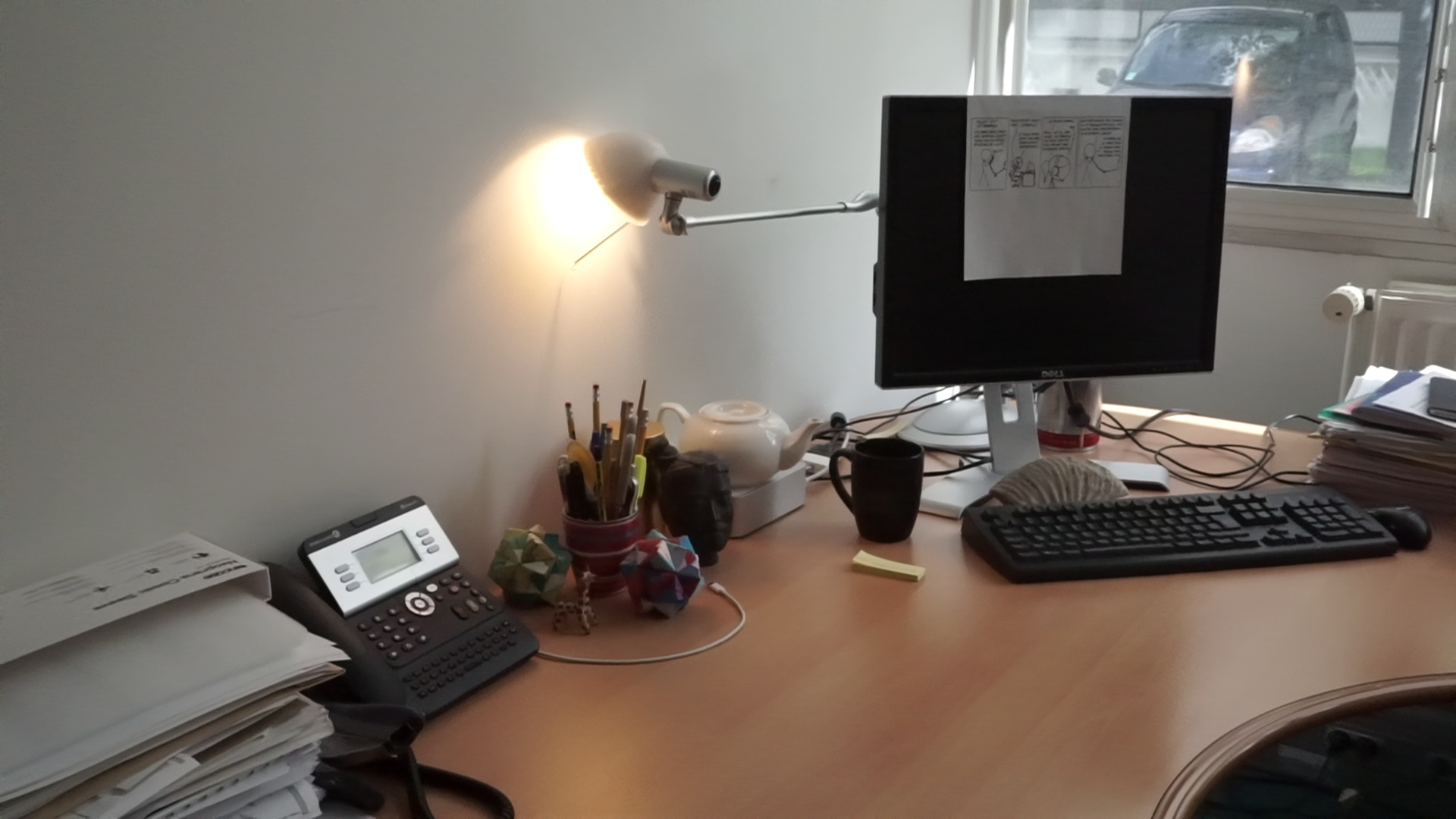}};
    \begin{scope}[x={(image.north east)},y={(image.south west)}]
        \draw[other2b, very thick] (0.35,0.6) rectangle (0.45,0.8); 
    \end{scope}
\end{tikzpicture}

\end{minipage}
\vspace{.3em}

\begin{minipage}[c]{\columnwidth}
\centering
\hspace{-1.2em}
\begin{sideways}
\hspace{3.8em}~Input 
\end{sideways}
\cfboxR{other1b}{\includegraphics[width=.305\textwidth]{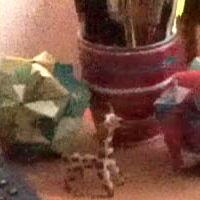}} 
\cfboxR{other1b}{\includegraphics[width=.305\textwidth]{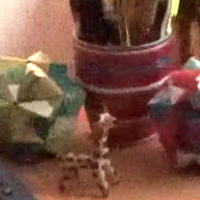}} 
\cfboxR{other1b}{\includegraphics[width=.305\textwidth]{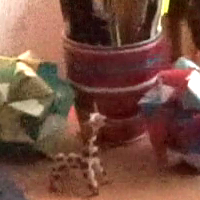}} 
\end{minipage}
\vspace{.3em}

\begin{minipage}[c]{\columnwidth}
\centering
\hspace{-1.2em}
\begin{sideways}
\hspace{3em} \textbf{Processed}
\end{sideways}
\cfboxR{other2b}{\includegraphics[width=.305\textwidth]{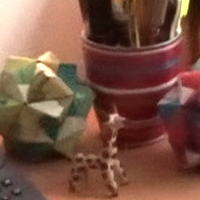}} 
\cfboxR{other2b}{\includegraphics[width=.305\textwidth]{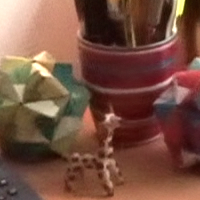}} 
\cfboxR{other2b}{\includegraphics[width=.305\textwidth]{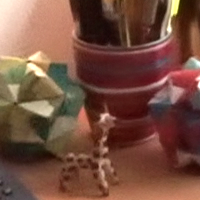}} 
\end{minipage}
\vspace{.3em}


{\footnotesize (a)}

\vspace{1em}


\begin{minipage}[c]{.49\columnwidth}
\centering

Input (sharp) frame
\vspace{.1em}

\begin{tikzpicture}
    \node[anchor=north west,inner sep=0] (image) at (0,0) {\includegraphics[width=\textwidth]{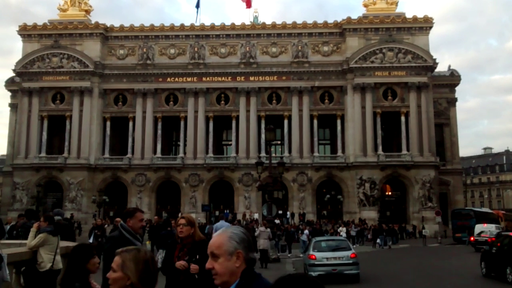}};
    \begin{scope}[x={(image.north east)},y={(image.south west)}]
        \draw[other1b, very thick] (0.095,0.7) rectangle (0.27,0.99); 
    \end{scope}
\end{tikzpicture}
\end{minipage}
\hspace{-.5em}
\begin{minipage}[c]{.49\columnwidth}
\centering

Processed frame
\vspace{.3em}

\begin{tikzpicture}
    \node[anchor=north west,inner sep=0] (image) at (0,0) {\includegraphics[width=\textwidth]{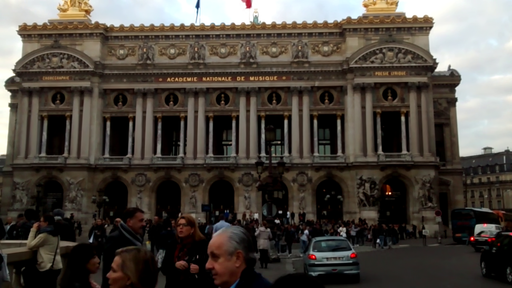}};
    \begin{scope}[x={(image.north east)},y={(image.south west)}]
        \draw[other2b, very thick] (0.095,0.7) rectangle (0.27,0.99); 
    \end{scope}
\end{tikzpicture}

\end{minipage}
\vspace{.3em}

\begin{minipage}[c]{\columnwidth}
\centering
\hspace{-1.2em}
\begin{sideways}
\hspace{2.5em}~Input 
\end{sideways}
\cfboxR{other1b}{\includegraphics[width=.2315\textwidth]{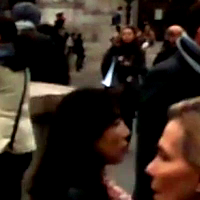}} 
\cfboxR{other1b}{\includegraphics[width=.2315\textwidth]{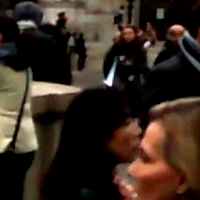}} 
\cfboxR{other1b}{\includegraphics[width=.2315\textwidth]{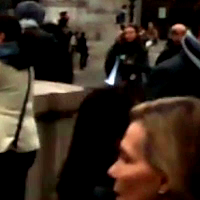}} 
\cfboxR{other1b}{\includegraphics[width=.2315\textwidth]{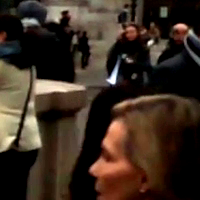}} 
\end{minipage}
\vspace{.3em}

\begin{minipage}[c]{\columnwidth}
\centering
\hspace{-1.2em}
\begin{sideways}
\hspace{1.4em} \textbf{Processed} 
\end{sideways}
\cfboxR{other2b}{\includegraphics[width=.2315\textwidth]{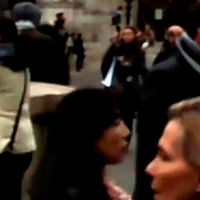}}
\cfboxR{other2b}{\includegraphics[width=.2315\textwidth]{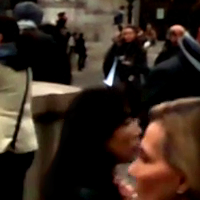}} 
\cfboxR{other2b}{\includegraphics[width=.2315\textwidth]{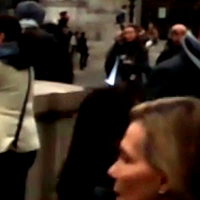}} 
\cfboxR{other2b}{\includegraphics[width=.2315\textwidth]{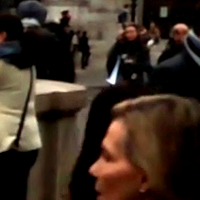}} 
\end{minipage}
\vspace{.3em}

{\footnotesize (b)}

\caption{ Processing already sharp sequences. When the input sequence is already sharp, the algorithm averages the input frames to reduce noise. This is illustrated by the first example (a) where three input frames crop and the respective restored versions are shown. 
In general, object occlusions are handled correctly by the consistent registration check as shown in the example (b). The full images and more results
regarding processed sequences that were already sharp are given in the supplementary material.}
\label{fig:sharp}

\end{figure}

\noindent \textbf{Noise reduction.}
A side effect of the proposed method is the reduction of video noise. Since the algorithm averages different frames, having different noise realizations, the final sequence will have less noise. This is shown in figures~\ref{fig:iterativeNoise} and~\ref{fig:sharp}, where from both a simple visual inspection and a quantitative analysis, it becomes clear that the noise is significantly reduced, in particular in the first pass of the algorithm. To that aim, we computed the level of noise in the images at each iteration, using the algorithm of~\cite{colom2013analysis} (see caption of Figure~\ref{fig:iterativeNoise} for details).

\vspace{.3em}

\noindent \textbf{Processing sharp sequences.} Typical videos target dynamic scenes with many objects moving in different directions and therefore there are potentially many occlusions.
Figure~\ref{fig:sharp} (b) shows an example of an already sharp sequence that was processed by the algorithm.  The consistency check  prevents the algorithm from averaging different parts (notably those that have been occluded and cannot be registered to nearby frames).

\vspace{.3em}

\begin{figure}[ptb]
\footnotesize
\centering

\begin{minipage}[c]{.49\columnwidth}
\centering

Input (blurry) frame
\vspace{.1em}

\begin{tikzpicture}
    \node[anchor=north west,inner sep=0] (image) at (0,0) {\includegraphics[width=\textwidth]{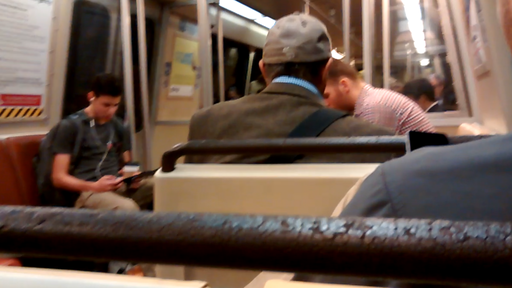}};
    \begin{scope}[x={(image.north east)},y={(image.south west)}]
        \draw[other1b, very thick] (0.6328,0.1389) rectangle (0.7891,0.4167); 
        \draw[red11b, very thick] (0.3984,0) rectangle (0.5547, 0.2778); 
        \draw[other7b, very thick] (0.8438,0.2917) rectangle (1.000,0.5694); 
    \end{scope}
\end{tikzpicture}
\end{minipage}
\hspace{-.5em}
\begin{minipage}[c]{.49\columnwidth}
\centering

Processed frame
\vspace{.3em}

\begin{tikzpicture}
    \node[anchor=north west,inner sep=0] (image) at (0,0) {\includegraphics[width=\textwidth]{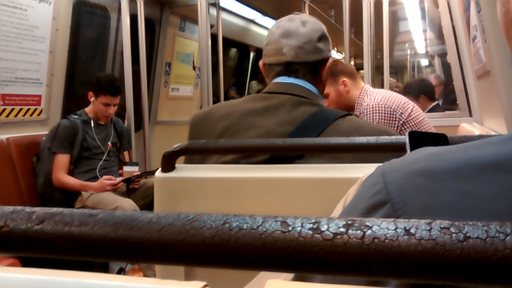}};
    \begin{scope}[x={(image.north east)},y={(image.south west)}]
        \draw[other2b, very thick] (0.6328,0.1389) rectangle (0.7891,0.4167); 
        \draw[red11a, very thick] (0.3984,0) rectangle (0.5547, 0.2778); 
        \draw[other6b, very thick] (0.8438,0.2917) rectangle (1.000,0.5694); 
    \end{scope}
\end{tikzpicture}

\end{minipage}


\vspace{.3em}


\begin{minipage}[c]{\columnwidth}
\centering
\hspace{-1.2em}
\begin{sideways}
\hspace{3.8em}~Input 
\end{sideways}
\cfboxR{red11b}{\includegraphics[width=.3125\textwidth]{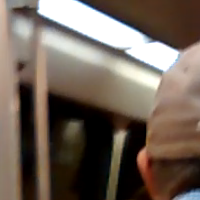}} 
\cfboxR{other1b}{\includegraphics[width=.3125\textwidth]{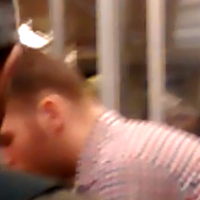}}
\cfboxR{other7b}{\includegraphics[width=.3125\textwidth]{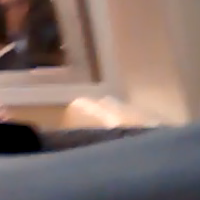}} 
\end{minipage}
\vspace{.3em}

\begin{minipage}[c]{\columnwidth}
\centering
\hspace{-1.2em}
\begin{sideways}
\hspace{3em}~\textbf{Processed}
\end{sideways}
\cfboxR{red11a}{\includegraphics[width=.3125\textwidth]{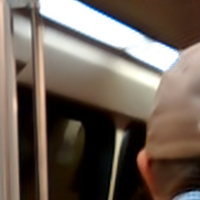}} 
\cfboxR{other2b}{\includegraphics[width=.3125\textwidth]{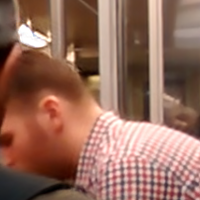}}
\cfboxR{other6b}{\includegraphics[width=.3125\textwidth]{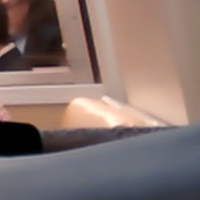}} 
\end{minipage}
\vspace{.1em}

{\footnotesize (a)}

\vspace{.4em}

\begin{minipage}[c]{\columnwidth}
\centering
\hspace{-1.2em}
\begin{sideways}
\hspace{2.0em}~Input 
\end{sideways}
\cfboxR{red11b}{\includegraphics[width=.2315\textwidth]{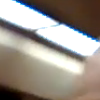}}
\cfboxR{red11b}{\includegraphics[width=.2315\textwidth]{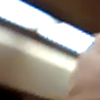}}
\cfboxR{red11b}{\includegraphics[width=.2315\textwidth]{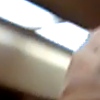}} 
\cfboxR{red11b}{\includegraphics[width=.2315\textwidth]{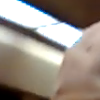}} 
\end{minipage}

\vspace{.3em}

\begin{minipage}[c]{\columnwidth}
\centering
\hspace{-1.2em}
\begin{sideways}
\hspace{1.3em}~\textbf{Processed}
\end{sideways}
\cfboxR{red11a}{\includegraphics[width=.2315\textwidth]{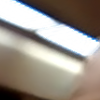}}
\cfboxR{red11a}{\includegraphics[width=.2315\textwidth]{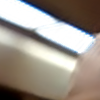}}
\cfboxR{red11a}{\includegraphics[width=.2315\textwidth]{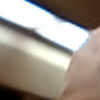}} 
\cfboxR{red11a}{\includegraphics[width=.2315\textwidth]{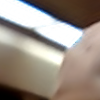}} 
\end{minipage}

\vspace{.4em}

{\footnotesize (b)}

\caption{Dealing with saturated regions. Saturated regions violate the linear convolution model. This presents a challenge for both image registration and deblurring. As these examples show, these regions are generally well processed by the proposed algorithm. Figure (a) shows some extracts containing saturated regions while (b) shows an extract of the left image crop for four consecutive frames. Frames differently blurred may cause differences in the saturated region size, this is not tackled by the algorithm.   More results
showing saturated regions are given in the supplementary material. }
\label{fig:saturated}

\end{figure}

\noindent \textbf{Dealing with saturated regions.} Videos present saturated regions in many situations. In these regions, the linear convolution model (blurring) is violated, presenting a challenge for both image registration and deblurring.  Figure~\ref{fig:saturated} shows different extracts of the {\small \textsf{metro}} sequence that present saturated regions.  In particular, saturated regions in blurry frames may change their size from one frame to another due to the difference in the respective frame blur. Thus, when registering these frames, depending on the size of the saturated region, the registration (which is based on an optical flow estimation) might find a non-rigid geometric transformation that puts into perfect correspondence these two regions (as they have the same color). In this case, the algorithm will not do any blur removal since all the frames have the same content (Figure~\ref{fig:saturated} (b) left crop). 
On the other hand, small saturated regions (like the green light in Figure~1,  the small light shown in Figure~\ref{fig:saturated} (b) middle crop, and the light reflection in Figure~\ref{fig:saturated} (b) right crop) are successfully deblurred since in this case the saturated region being very small,  the registration algorithm  rigidly transfers a sharp version found in a nearby frame. 
The compromise between these two behaviors is given by the optical flow estimation algorithm.
In general, the algorithm successfully handles these cases.  More results showing saturated regions are given in the supplementary material.

\vspace{.3em}

\vspace{.3em}

\begin{figure}
\footnotesize
\centering

\begin{minipage}[c]{.48\columnwidth}
\centering

\includegraphics[width=\textwidth]{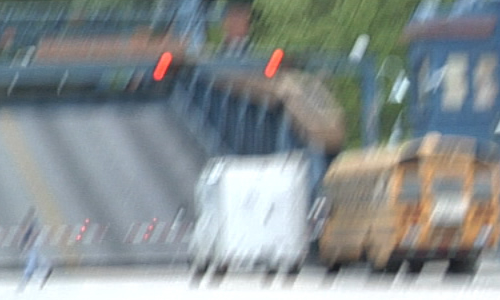} \vspace{-.9em}

\includegraphics[width=\textwidth]{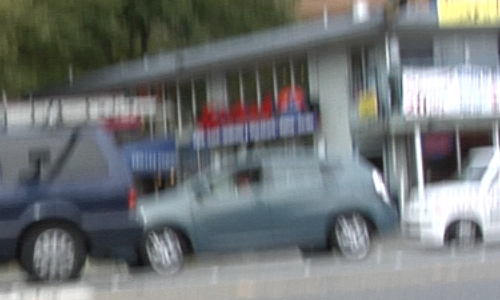}

Blurry input

\end{minipage}
\begin{minipage}[c]{.48\columnwidth}
\centering

\includegraphics[width=\textwidth]{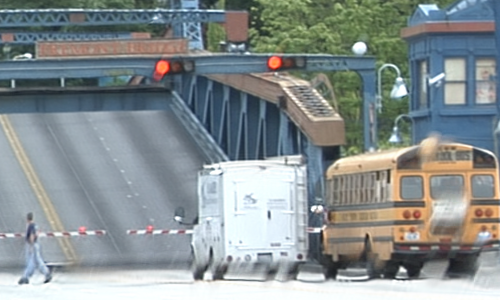} \vspace{-.9em}

\includegraphics[width=\textwidth]{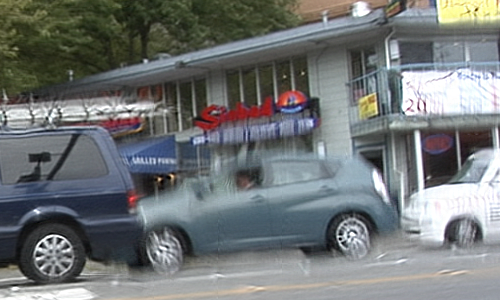}

\textbf{Proposed method}

\end{minipage}

\caption{Partial failure cases, unrealistic \emph{rendering}. During extreme camera shake, correctly registering the input frames is a very challenging task. The consistent registration may lead to image regions that are not deblurred, and thus create some visual unrealistic artifacts, as in parts of the yellow bus. Although the bus is reconstructed mostly sharp, there are some still blurry parts. In the sequence shown on the bottom, one would accept to see the car blurred in the driving direction. However, in this case, some of the car's blur is in the vertical direction since the blurring is coming  from the vertical motion of the camera and not from the car.}
\label{fig:failure1}

\end{figure}

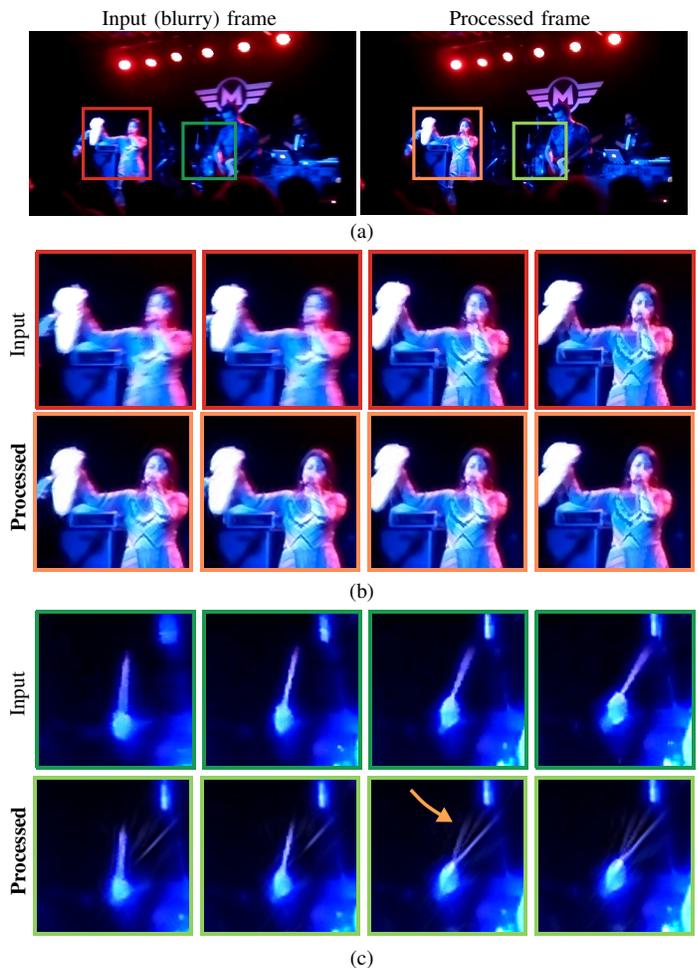
\begin{figure}
\footnotesize
\centering
\begin{minipage}[c]{.49\columnwidth}
\centering

Input (blurry) frame
\vspace{.1em}

\begin{tikzpicture}
    \node[anchor=north west,inner sep=0] (image) at (0,0) {\includegraphics[width=\textwidth]{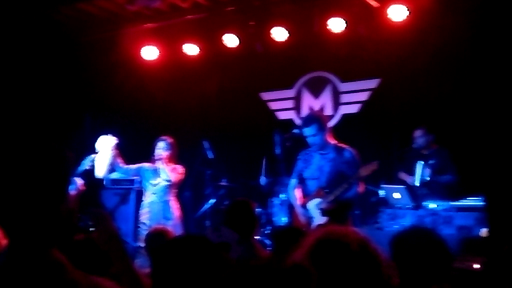}};
    \begin{scope}[x={(image.north east)},y={(image.south west)}]
        \draw[other1b, very thick] (0.165,0.42) rectangle (0.37,0.8); 
        \draw[other7b, very thick] (0.47,0.5) rectangle (0.63,0.8); 
    \end{scope}
\end{tikzpicture}
\end{minipage}
\hspace{-.5em}
\begin{minipage}[c]{.49\columnwidth}
\centering

Processed frame
\vspace{.3em}

\begin{tikzpicture}
    \node[anchor=north west,inner sep=0] (image) at (0,0) {\includegraphics[width=\textwidth]{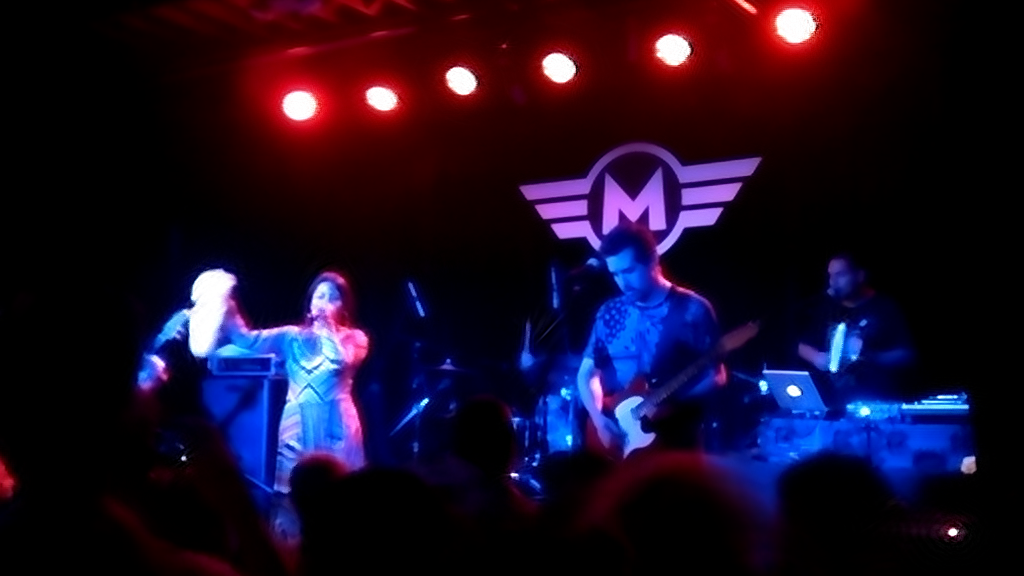}};
    \begin{scope}[x={(image.north east)},y={(image.south west)}]
        \draw[other2b, very thick] (0.165,0.42) rectangle (0.37,0.8); 
        \draw[other6b, very thick] (0.47,0.5) rectangle (0.63,0.8); 
    \end{scope}
\end{tikzpicture}

\end{minipage}
\vspace{.4em}

(a)

\vspace{.4em}

\begin{minipage}[c]{\columnwidth}
\centering
\hspace{-1.2em}
\begin{sideways}
\hspace{2.0em}~Input 
\end{sideways}
\cfboxRR{other1b}{\includegraphics[width=.23\textwidth]{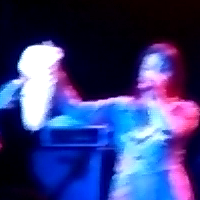}} 
\cfboxRR{other1b}{\includegraphics[width=.23\textwidth]{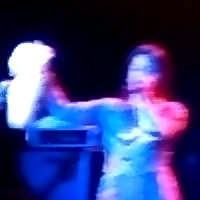}} 
\cfboxRR{other1b}{\includegraphics[width=.23\textwidth]{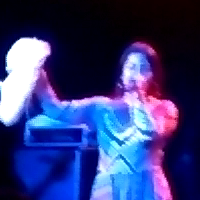}} 
\cfboxRR{other1b}{\includegraphics[width=.23\textwidth]{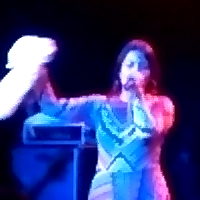}} 
\end{minipage}
\vspace{.1em}

\begin{minipage}[c]{\columnwidth}
\centering
\hspace{-1.2em}
\begin{sideways}
\hspace{1.4em}~\textbf{Processed} 
\end{sideways}
 \cfboxRR{other2b}{\includegraphics[width=.23\textwidth]{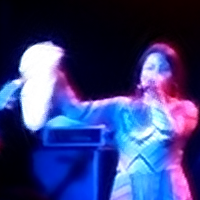}} 
 \cfboxRR{other2b}{\includegraphics[width=.23\textwidth]{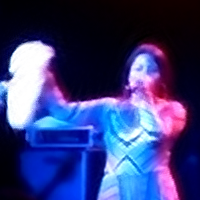}} 
 \cfboxRR{other2b}{\includegraphics[width=.23\textwidth]{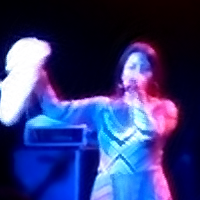}} 
 \cfboxRR{other2b}{\includegraphics[width=.23\textwidth]{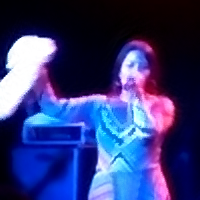}} 
\end{minipage}
\vspace{.1em}

(b)

\vspace{.4em}

\begin{minipage}[c]{\columnwidth}
\centering
\hspace{-1.2em}
\begin{sideways}
\hspace{2.0em}~Input 
\end{sideways}
\cfboxRR{other7b}{\includegraphics[width=.23\textwidth]{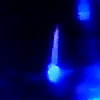}}  
\cfboxRR{other7b}{\includegraphics[width=.23\textwidth]{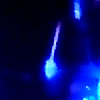}}  
\cfboxRR{other7b}{\includegraphics[width=.23\textwidth]{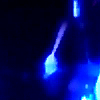}}  
\cfboxRR{other7b}{\includegraphics[width=.23\textwidth]{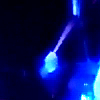}} 
\end{minipage}
\vspace{.3em}

\begin{minipage}[c]{\columnwidth}
\centering
\hspace{-1.2em}
\begin{sideways}
\hspace{1.4em}~\textbf{Processed} 
\end{sideways}
\cfboxRR{other6b}{ \includegraphics[width=.23\textwidth]{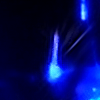}} 
\cfboxRR{other6b}{ \includegraphics[width=.23\textwidth]{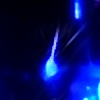}} 
\cfboxRR{other6b}{\includegraphics[width=.23\textwidth]{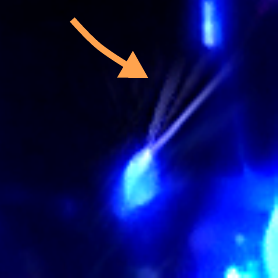}} 
\cfboxRR{other6b}{\includegraphics[width=.23\textwidth]{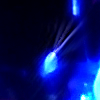}}

\vspace{.4em}

(c)
\end{minipage}

\caption{Partial failure cases due to miss-registration of small details. Figure (a) shows an input blurry frame from {\scriptsize \textsf{anita}} sequence (left) and the respective frame processed by the proposed method (right). On the top row of (b) and (c), two image crops from four successive input frames are shown, while the second rows of (b) and (c) show the same image crops extracted from the respective processed frames. In general, as shown in (b) the algorithm correctly manages to deblur the blurry regions. Nevertheless, in some very small details the frame registration (based on the optical flow estimation) might fail and lead to the introduction of minor artifacts. An example of this is shown in (c) in the drummer's percussion mallet (pointed by the arrow). This does not happen often as shown in the supplementary video.}
\label{fig:failure2}

\end{figure}

\noindent \textbf{Partial failure cases.}
When the blur is extreme, correctly registering the input frames is very challenging. In some of these difficult cases, our consistent registration may lead to image regions that are not sufficiently deblurred. This creates visual artifacts, as in the yellow bus in Figure~\ref{fig:failure1}. Although the bus is mostly sharp in the restored frame, it contains some blurry parts. Also, very small and not contrasted details can be very difficult to register with the considered approach. 
This is due to an intrinsic ambiguity on the optical flow computation introduced by the blur.  This may introduce small artifacts as shown in Figure~\ref{fig:failure2}. Despite these particular local cases, the proposed algorithm produces images of very good quality.

\section{Discussion, Limitations and Future Work}

Videos captured with hand-held cameras often present blurry frames due to the camera shake. In this work, we have presented an algorithm that addresses this particular deblurring scenario. The proposed method relies on the fact that, in a blurry video, frames are generally differently blurred, as a consequence of the random nature of hand tremor.

The proposed method is based on the Fourier Burst Accumulation principle. By computing a weighted average in the Fourier domain, we reconstruct an image combining the least attenuated frequencies in each frame. The proposed algorithm is not a universal deblurring algorithm in the sense that it assumes that the frames are differently blurred. In particular, the proposed method will not handle the case of blur due to a camera panning at constant speed. 

The key idea of the proposed algorithm is to consistently register nearby frames to each frame in the input sequence. This avoids artifacts in the frames fusion. Similar ideas have been explored before, but the efforts have been focused on trying to find similar patches in nearby frames. Here, we concentrate on creating a new compatible set of consistent frames that allows a local weighted Fourier fusion. Moreover, since the algorithm introduces very limited artifacts, it can be iterated to propagate the fusion information to farther frames without the risk of introducing noticeable damage.

Another important aspect of the proposed approach, is that it does not degrade the quality of originally good sharp frames. If the only sharp frame in the set is the reference, the Fourier weighting scheme will automatically select this one.  While, if there are several sharp frames on top of the reference, the consistent registration procedure will avoid degrading the quality (e.g., ghosting artifacts).

Extensive experimental results showed that the algorithm is fast and easy to implement. As a future work, we would like to explore other possible ways of computing the consistent optical flow, since this is the current computational bottleneck. Also, we would like to handle the failure cases due to a wrong registration. For that end, one possible venue is to explore other occlusions detection algorithms  using more than two consecutive frames (as done in~\cite{ince2008occlusion,ballester2012}). However, this is very challenging if we want to keep the algorithmic complexity low.

\ifCLASSOPTIONcaptionsoff
  \newpage
\fi

\bibliographystyle{IEEEtran}
\bibliography{burst_deblurring_clean_short_video2015_clean}

\begin{IEEEbiography}[{\includegraphics[width=1in,height=1.25in,clip,keepaspectratio]{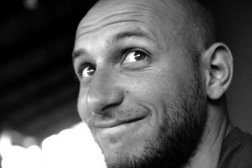} \vspace{4em}}]{Mauricio Delbracio}
received the graduate degree from the Universidad
de la Rep\'{u}blica, Uruguay, in electrical engineering in 2006, the MSc
and PhD degrees in applied mathematics from \'{E}cole normale sup\'{e}rieure de Cachan, France, in 2009
and 2013 respectively. He currently has a postdoctoral position at
the Department of Electrical and Computer Engineering, Duke
University. His research interests include image and signal
processing, computer graphics, photography and computational
imaging.
\end{IEEEbiography}
%
\begin{IEEEbiography}[{\includegraphics[width=1in,height=1.25in,clip,keepaspectratio]{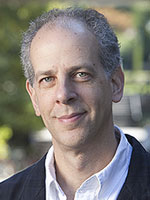}}]{Guillermo Sapiro}

(F'14) was born in Montevideo, Uruguay, on April 3, 1966. He received his B.Sc. (summa cum laude), M.Sc., and Ph.D. from the Department of Electrical Engineering at the Technion, Israel Institute of Technology, in 1989, 1991, and 1993 respectively. After post-doctoral research at MIT, Dr. Sapiro became Member of Technical Staff at the research facilities of HP Labs in Palo Alto, California. He was with the Department of Electrical and Computer Engineering at the University of Minnesota, where he held the position of Distinguished McKnight University Professor and Vincentine Hermes-Luh Chair in Electrical and Computer Engineering. Currently he is the Edmund T. Pratt, Jr. School Professor with Duke University.

G. Sapiro works on theory and applications in computer vision, computer graphics, medical imaging, image analysis, and machine learning. He has authored and co-authored over 300 papers in these areas and has written a book published by Cambridge University Press, January 2001.

G. Sapiro was awarded the Gutwirth Scholarship for Special Excellence in Graduate Studies in 1991, 
the  Ollendorff Fellowship for Excellence in Vision and Image Understanding Work in 1992, 
the Rothschild Fellowship for Post-Doctoral Studies in 1993, the Office of Naval Research Young Investigator Award in 1998, 
the Presidential Early Career Awards for Scientist and Engineers (PECASE) in 1998, the National Science Foundation Career Award in 1999, and
the National Security Science and Engineering Faculty Fellowship in 2010.
He received the test of time award at ICCV 2011.

G. Sapiro is a Fellow of IEEE and SIAM.

G. Sapiro was the founding Editor-in-Chief of the SIAM Journal on Imaging Sciences.

\end{IEEEbiography}

\vfill

\end{document}